\documentclass[10pt,journal,compsoc]{IEEEtran}
\ifCLASSOPTIONcompsoc
  \usepackage[nocompress]{cite}
\else
  \usepackage{cite}
\fi
\ifCLASSINFOpdf
\else
\fi
\usepackage{cite}
\usepackage{amsmath,amssymb,amsfonts}
\usepackage{graphicx}
\usepackage{textcomp}
\usepackage{xcolor}
\usepackage{microtype}
\usepackage{graphicx}
\usepackage{subfigure}
\usepackage{booktabs} 
\usepackage{multirow}
\usepackage{xcolor}
\usepackage{amsmath}
\usepackage{amssymb}
\usepackage{algpseudocode}
\usepackage{algorithm}
\usepackage{forloop}

\algnewcommand\And{\textbf{and}}
\algnewcommand\Or{\textbf{or}}
\DeclareMathOperator*{\argmax}{arg\,max}
\def\BibTeX{{\rm B\kern-.05em{\sc i\kern-.025em b}\kern-.08em
    T\kern-.1667em\lower.7ex\hbox{E}\kern-.125emX}}
\usepackage{pifont}

\begin{document}

\title{Scalable Pathogen Detection from Next Generation DNA Sequencing with Deep Learning}

\author{Sai Narayanan, Sathyanarayanan Aakur$^*$,~\IEEEmembership{Member,~IEEE,}
        Priyadharsini Ramamurthy, Arunkumar Bagavathi, Vishalini Ramnath and Akhilesh Ramachandran$^*$
\IEEEcompsocitemizethanks{\IEEEcompsocthanksitem 
$^*$ denotes corresponding author. E-mail: \{saakurn,rakhile\}@okstate.edu \\SN Aakur, P. Ramamurthy, A. Bagavathi and V. Ramnath are with the Department of Computer Science, Oklahoma State University, Stillwater, OK, 74078.\protect\\
\IEEEcompsocthanksitem S. Narayanan and A. Ramachandran are with Oklahoma Animal Disease Diagnostic Laboratory, Oklahoma State University, Stillwater, OK, 74078.}
\thanks{Manuscript received April 19, 2005; revised August 26, 2015.}}

%
%

\markboth{Journal of \LaTeX\ Class Files,~Vol.~14, No.~8, August~2015}%
{Shell \MakeLowercase{\textit{et al.}}: Bare Demo of IEEEtran.cls for Computer Society Journals}
%



\IEEEtitleabstractindextext{%
\begin{abstract}
Next-generation sequencing technologies have enhanced the scope of Internet-of-Things (IoT) to include genomics for personalized medicine through the increased availability of an abundance of genome data collected from heterogeneous sources at a reduced cost. Given the sheer magnitude of the collected data and the significant challenges offered by the presence of highly similar genomic structure across species, there is a need for robust, scalable analysis platforms to extract actionable knowledge such as the presence of potentially zoonotic pathogens. The emergence of zoonotic diseases from novel pathogens, such as the influenza virus in 1918 and SARS-CoV-2 in 2019 that can jump species barriers and lead to pandemic underscores the need for scalable metagenome analysis. In this work, we propose \textit{MG2Vec}, a deep learning-based solution that uses the transformer network as its backbone, to learn robust features from raw metagenome sequences for downstream biomedical tasks such as targeted and generalized pathogen detection. Extensive experiments on four increasingly challenging, yet realistic diagnostic settings, show that the proposed approach can help detect pathogens from uncurated, real-world clinical samples with minimal human supervision in the form of labels. Further, we demonstrate that the learned representations can generalize to completely unrelated pathogens across diseases and species for large-scale metagenome analysis. We provide a comprehensive evaluation of a novel representation learning framework for metagenome-based disease diagnostics with deep learning and provide a way forward for extracting and using robust vector representations from low-cost next generation sequencing to develop generalizable diagnostic tools. 
\end{abstract}

}

\maketitle

\IEEEdisplaynontitleabstractindextext

%
\IEEEpeerreviewmaketitle

\IEEEraisesectionheading{\section{Introduction}\label{sec:introduction}}

The Internet-of-Things (IoT) ecosystem has rapidly evolved and has provided an exciting new dimension for advances in technologies for healthcare. Combined with advances in DNA sequencing technologies~\cite{metzker2010sequencing,mikheyev2014first}, the world of bioinformatics has significant strides towards potential applications in precision medicine and personalized healthcare. For example, portable sequencing models are already showing great potential in genetic biomonitoring~\cite{krehenwinkel2019genetic,pomerantz2018real}, conservation~\cite{blanco2019next} and even plant pathogen detection~\cite{bronzato2018nanopore}. 
The sheer magnitude, complexity, and diversity of the collected data about biological components, processes, and systems are represented by DNA sequences from environmental and clinical samples called \textit{metagenomes}. This data contains vital information about both the host (the human, animal or plant from which the sample was collected) as well as the foreign entities (such as bacteria, fungi or viruses) that could be present in the obtained sample. The DNA sequences can provide significant amounts of actionable information about both parties that can be used for diagnostics and personalized healthcare. It can then be used to derive information such as the nature of the foreign agent including its pathogenic potential and antimicrobial resistance as well as the potential response of the host to the drugs that could be used to treat any diseases caused by the foreign agents. 

DNA sequencing-based solutions hold significant potential for disease diagnostic applications. Multiple sequencing technologies such as the sequencing by synthesis technology (Illumina) and single molecule nanopore based sequencing (Oxford Nanopore Technologies - ONT) has made sequencers easy to use and portable in many areas. The portable nanopore sequencer (MinION) has been used to identify Ebola virus~\cite{yong2015fighting,hoenen2016nanopore} and other pathogens~\cite{dippenaar2021nanopore,brown2017minion,monzer2020prevalence}. Metagenome sequencing of clinical samples for pathogen detection can be grouped under two classes viz., targeted sequencing and shotgun sequencing. Targeted sequencing involves the use of Polymerase Chain Reaction (PCR)-based amplification of target regions in bacteria or fungi, following which the PCR amplified regions can be sequenced to identify different organisms. Shotgun sequencing is the process of sequencing the entire sample directly without any targeted PCR amplification. Unlike targeted sequencing, possibilities for introducing taxonomic bias during sample preparation for sequencing is reduced in shotgun sequencing.

\begin{table}[t]
    \centering
    \caption{\textbf{Genome Similarity.} The many pathogens that belong to the same or closely related families can have highly similar genomes and hence overlapping sequences. There is a need for fine-grained recognition to distinguish between them.}
    
    \resizebox{0.99\columnwidth}{!}{
    \begin{tabular}{|c|c|c|c|c|c|c|}
    \toprule
   \textbf{Pathogen}  & \textbf{H. som.} & \textbf{M. bov.}  & \textbf{M. hae.}  & \textbf{P. mul.} & \textbf{T. pyo.} & \textbf{B. Tre.} \\ \toprule
    \textbf{H. somni} & - & 0.797 & 0.876 & \textbf{0.884} & 0.794 &\textbf{ 0.847} \\ \midrule
    \textbf{M. bovis} & 0.797 & - & 0.794 & \textbf{0.801} & \textbf{0.806} & \textbf{0.800} \\ \midrule
    \textbf{M. haemo.} & 0.876 & 0.794 & - & \textbf{0.955} & 0.792 & \textbf{0.945} \\ \midrule
    \textbf{P. multoc.} & \textbf{0.884} & \textbf{0.801} & \textbf{0.955} & - & 0.786 & \textbf{0.847} \\ \midrule
    \textbf{T. pyoge.} & 0.794 & \textbf{0.806} & 0.792 & 0.786 & - & 0.789 \\ \midrule
    \textbf{B. trehal.} & \textbf{0.847} & \textbf{0.800} &\textbf{ 0.945} & \textbf{0.847} & 0.789 & - \\ \midrule
    \end{tabular}
    }
    \label{tab:genome_sim}
\end{table}

\textbf{What are metagenomes?} A metagenome is a collection of DNA sequences that are extracted from clinical or environmental samples, that contain nucleic acid from multiple sources. The key difference between a genome and a metagenome is that the genome comprises the genetic information of one particular species whereas metagenomes contains a mixture of DNA from multiple species sourced from all cells in a community. This data offers some unique opportunities for biomedical and healthcare applications. Instead of targeting a specific genomic locus, as in the case of 16S rRNA (where a specific genomic marker is targeted and amplified), the entire DNA content from the sample is sheared into individual segments or ``\textit{reads}'' and independently sequenced in a shotgun approach. The resulting reads represent various genomic locations for all genomes present in the sample, including non-microbes. Hence, they are not confined or targeted towards specific species, but can offer possible infinite multiplexing, i.e., the ability to detect any species of interest from DNA sequence reads without bias. Traditional bioinformatics-based approaches~\cite{johnson2008ncbi,wood2014kraken,wood2019improved} have enabled great science; however, they appear to be hitting the wall of efficiency and scalability. They are target specific and sometimes fail to decrypt complex relationships in the data to reveal new discoveries. There is a need for \textit{pathogen-agnostic} metagenome analysis frameworks that can help segment and cluster metagenome sequences into smaller and concentrated genome reads that belong to the same pathogen species without prior knowledge of disease etiologies, while preserving privacy~\cite{wang2016big,hu2017secure}. Such an approach will overcome the multiplexing limitations of traditional methods and help detect novel and emerging pathogens that are not routinely probed for. However, this requires a robust representation of the metagenome sequence that can distinguish between closely related species.

\textbf{Challenges.} Metagenome analysis comes with a set of challenges that we term as the {\textit{small} big data problem}. While metagenomes extracted from clinical samples can have millions of nucleotide sequence reads, many of them belong to the host with a few strands of pathogen sequences interspersed. This results in a long-tail distribution between the different genome sequences that can be present in the sample. This problem becomes more intense when we consider the similarity between the genome of pathogens that belong to the same family, which can be as high as $98\%$. For example, consider the Bovine Respiratory Disease Complex (BRD). This complex multi-etiologic disease affects cattle worldwide and is one of the leading causes of economic distress in the cattle industry. The chief bacterial pathogens are \textit{Mannheimia haemolytica}, \textit{Pasteurella multocida}, \textit{Bibersteinia trehalosi}, \textit{Histophilus somni}, \textit{Mycoplasma bovis}, and \textit{Trueperella pyogenes}, which belong to the \textit{Pasteurellaceae}, \textit{Actinomycetaceae} and \textit{Mycobacteriaceae} families. 
A comparison of the genome similarity between the common pathogens in BRDC is shown in Table~\ref{tab:genome_sim}. It can be seen that they are highly similar, with only \textit{T. Pyogenes} showing more variation. Considering that the average bacteria genome is around $3.7$ million bases, a similarity of $80\%$ between the genomes makes the pathogen detection task extremely hard, particularly considering that metagenome reads from clinical or environmental samples can have high amounts of noise due to observation error~\cite{laver2015assessing}. Note that this turns the pathogen identification problem to a fine-grained detection problem where the goal is to identify metagenome sequences unique to each species to definitively detect their presence. 

\begin{figure}[t]
    \centering
        \includegraphics[width=0.99\columnwidth]{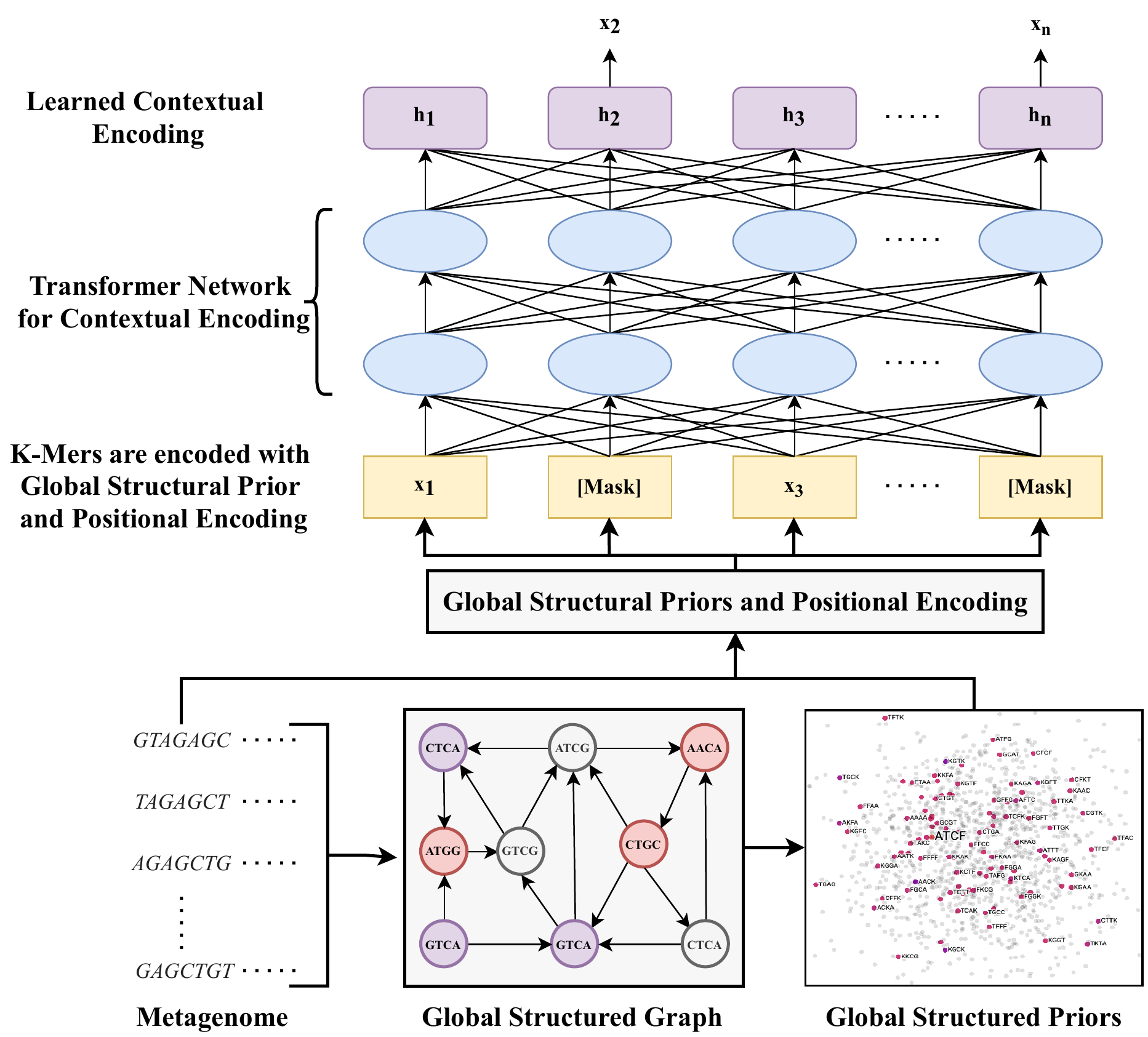}  \\
    \caption{The \textbf{overall approach} for learning robust representations is illustrated. Given an \textit{unlabeled} metagenome sequence, we learn a contextualized representation conditioned upon a global structural prior. The resulting representations can be used for scalable metagenome analysis with minimal labeled data.}
    \label{fig:arch}
\end{figure}

To address this challenging task, we aim to leverage advances in deep learning and computational biology to propose a scalable, metagenome analysis framework (\textit{MG2Vec}) to learn robust representations from DNA sequence reads. The goal is to learn robust representations from \textit{unlabeled} metagenome data that can capture fine-grained differences between highly similar pathogen samples. The overall approach is shown in Figure~\ref{fig:arch}. We explicitly capture global structural properties of metagenome sequences and contextualize them with local, sequence-level information to learn robust representations for scalable analysis. We show, through extensive experiments, that this self-supervised learning framework can help in pathogen detection in real-world clinical samples across four challenging settings. Specifically, we show that the framework can perform targeted pathogen detection in both curated and uncurated clinical metagenome samples with a high degree of precision using highly limited, unbalanced data with a long-tail distribution. We also show that the representations learned from the framework can generalize well to segment clinical metagenome sequences containing \textit{unseen} and \textit{unknown} pathogen infections with \textit{no supervision}. Finally, we demonstrate that the learned representations can generalize to completely unrelated pathogens across diseases and species for large-scale metagenome analysis. We provide a comprehensive evaluation of a novel representation learning framework for metagenome-based disease diagnostics with deep learning and provide a way forward for extracting and using robust vector representations from low-cost next generation sequencing to develop generalizable diagnostics. 

A preliminary version of this work appears as Metagenome2Vec~\cite{aakur2021metagenome2vec}, one of the first approaches to demonstrate the applicability of transformers towards pathogen detection from raw DNA sequence reads. The current paper is a significant extension of the work in terms of theory and extensive evaluation. First, we present a detailed introduction to the proposed framework and the necessary background information to understand the contributions of the framework. Second, we provide an extensive analysis of the applicability of the proposed framework to large-scale clinical data under a variety of constrained and unconstrained evaluation settings that assesses its generalization capabilities to derive insights from real-world data. We compare against machine learning baselines as well as traditional bioinformatics pipelines to provide a comprehensive evaluation of the proposed MG2Vec framework. 
We structure the paper as follows. In Section~\ref{related-work}, we provide an overview of prior, related works and connect them with the techniques used in this work. We provide the necessary background and the proposed framework in Section~\ref{sec:proposed}. In Section~\ref{sec:exp_setup}, we describe the experimental setup, including the evaluation settings and the metrics and baselines, used for assessing the performance of the proposed \textit{MG2Vec} framework. Section~\ref{sec:results} presents a thorough performance evaluation of the proposed approach along with ablation studies. Finally, in Section~\ref{sec:conclusion}, we conclude with a discussion on the framework and future directions of research that are opened up by the proposed approach. 

\section{Related Work}\label{related-work}

\textbf{Traditional metagenome analysis} has largely been tackled through bioinformatics approaches~\cite{olman2008parallel,liao2013new} such as BLAST~\cite{altschul1990basic}, Kraken~\cite{wood2014kraken,wood2019improved}, and Centrifuge~\cite{kim2016centrifuge}. BLAST is one of the most well-known and trusted approach for taxonomic classification of genome sequences. It uses a greedy algorithm to find the best alignment of the sequence to a species from a large search space. Kraken~\cite{wood2014kraken} and its successors~\cite{wood2019improved} are designed to be extremely lightweight for fast classification of metagenome sequences using exact alignment of k-mers from the query sequence to a database of pathogens. Similarly, Centrifuge employs the Burrows-Wheeler transform (BWT) and the Ferragina-Manzini (FM) index as part of an efficient indexing mechanism optimized specifically for the metagenomic classification problem. However, they appear to be hitting the wall of efficiency and scalability. They generally require computationally expensive alignment steps, can be target-specific, and sometimes fail to decrypt complex relationships in the data to reveal new discoveries. They are taxonomy dependent and follow exhaustive search on millions of metagenome sequence reads with a database of target sequence to identify pathogens. Such methods do not scale for large metagenomes and often do not generalize for novel pathogens since they look for sequence-level matches within a specific search space. 

\textbf{Machine learning-based metagenome analysis} have provided an exciting alternative~\cite{busia2019deep,rojas2019genet,mock2019viral,liang2020deepmicrobes,bartoszewicz2020deepac,queyrel2021towards,liu2020rnn,storato2021k2mem,goudarzi2017heterogeneous} to bioinformatics pipelines, largely driven by the impressive gains made by deep learning~\cite{lecun2015deep} models in language~\cite{vaswani2017attention,devlin2019bert} and vision~\cite{krizhevsky2012imagenet,He_2016_CVPR}. 
Deneke \textit{et al.}, in their seminal work, proposed PaPrBaG~\cite{deneke2017paprbag} where a random forest model was used for predicting the pathogenic potential of genome sequence reads, i.e., a binary classification task to determine whether the sequence read originated from pathogenic bacterial species or not and demonstrated its generalization capabilities towards classifying novel, unseen species. 
Busia \textit{et al.}~\cite{busia2019deep} proposed a deep learning model using depth-wise separable convolutional layers for read-level taxonomic classification from 16S ribosomal DNA sequences. 
Rojas \textit{et al.} proposed GeNet~\cite{rojas2019genet}, a deep convolutional neural network (CNN) for metagenomic classification from raw DNA sequences using standard backpropagation. 
Mock \textit{et al.}~\cite{mock2019viral} proposed the use of a class of recurrent neural networks (RNNs) called Long Short Term Memory networks (LSTMs)~\cite{hochreiter1997long} and a CNN-LSTM model for viral host prediction from next generation sequencing reads. 
Liang \textit{et al.} proposed DeepMicrobes~\cite{liang2020deepmicrobes}, a bi-directional LSTM model that uses attention to learn representations from k-mers derived from metagenome sequence reads. 
Bartoszewicz \textit{et al.} proposed DeePAC~\cite{bartoszewicz2020deepac}, a two-stream network of CNN-LSTMS that extracted representations from metagenome sequences and their reverse-complements. 
Queyrel \textit{et al.} proposed the closely related, and similarly named \textit{metagenome2vec}~\cite{queyrel2021towards}, for directly predicting diseases from raw metagenome sequences. They employ an attention mechanism to combine k-mer features from metagenome sequences, followed by a multiple instance learning classifier for end-to-end disease prediction. Narayanan \textit{et al.}~\cite{narayanan2020genome} and Indla \textit{et al.}~\cite{Indla2021Sim2RealFM} propose to use graph-based features from de Bruijn graphs constructed from metagenome sequences for species prediction from metagenome sequences, while Aakur \textit{et al.} presented MG-NET~\cite{aakur2021mg}, a multi-modal approach to pathogen detection from metagenome sequences using the idea of pseudo-images. Angelov~\cite{angelov2019species2vec} proposed \textit{species2vec} a random forest classifier that used embeddings learned from genome sequences for species identification. In addition to diagnosis, metagenome analysis has several applications in areas such as microbial taxonomic profiling~\cite{vervier2016large,nalbantoglu2018mimosa}, bacteria detection~\cite{fiannaca2018deep}, splice junction prediction~\cite{lee2015boosted}, dna-protein binding prediction~\cite{shadab2020deepdbp}, and clustering single-cell gene expressions~\cite{prabhakaran2016dirichlet}, to name a few.

\textbf{Self-supervised representation learning} approaches~\cite{vaswani2017attention,devlin2019bert,chen2022predicting,nguyen2021gefa} have led to tremendous progress in learning robust representations from unlabeled data i.e., raw data without human-supervised labels. Using a ``pretext'' task during the pre-training phase, these approaches learn supervisory signals directly from the data without the need for explicit supervision. 
The typical pipeline for such approaches is to first collate large amounts of unlabeled data in the relevant modality i.e., vision or language. Second, they design one or more pre-training objective functions that cleverly exploit the dependencies between entities within the data to provide supervision for learning general, but relevant representations from the unlabeled data. After a long pre-training stage, these models are then transfer-learned into specific tasks using smaller amounts of annotated training data. The goal is to learn generalized, transferable representations from big data that can potentially be useful under more constrained scenarios. 
They have shown incredible improvements in models in many modalities such as vision, text and audio, to name a few. 
In computer vision, self-supervised learning has leveraged the complex nature of visual signals to design pretext tasks for downstream tasks such as event segmentation~\cite{aakur2019perceptual}, anomaly detection~\cite{li2021cutpaste}, object detection~\cite{lee2019multi} and multimodal understanding~\cite{lu2019vilbert}, to name a few. In text, transformer networks~\cite{vaswani2017attention} have become the \textit{de facto} backbone networks for ``large language models'' or LLMs such as BERT~\cite{devlin2019bert} and GPT~\cite{radford2018improving,radford2019language,brown2020language}, which have shown tremendous performance in downstream tasks such as question answering~\cite{mihaylov2018can}. 

In this paper, we build upon our prior work~\cite{aakur2021metagenome2vec}, to propose MG2Vec, a transformer-based framework that proposes to train transformer network~\cite{vaswani2017attention} for species-agnostic, self-supervised pre-training from raw metagenome sequences for both targeted and generalized pathogen detection. We differ from the existing work in three key ways. First, we use uncurated, ``\textit{in-the-wild}'' clinical metagenome sequences instead of sampled, simulated metagenome data. We do not make any prior assumptions on the species of interest by employing 16S ribosomal DNA sequences or assembly as a pre-processing step. Second, we deal with a real-world scenario with several closely related pathogens with significantly fewer observed sequences, leading to a highly unbalanced training dataset with limited labels. Third, we provide four (4) evaluation settings with both targeted and generalized pathogen detection from uncurated metagenome sequences with significant amount of challenges such as distractor sequences, highly similar pathogens of interest and data with a significant long tail distribution.
%
\section{The Proposed MG2Vec Framework}\label{sec:proposed}

In this section, we present the proposed MG2Vec framework for constructing contextualized repersentations from metagenome sequences for scalable pathogen detection from next generation sequencing approaches. The overall approach is shown in Figure~\ref{fig:arch}. The goal is to learn robust representations from \textit{unlabeled} metagenome data that can capture fine-grained differences between highly similar pathogen samples. We explicitly capture global structural properties of metagenome sequences and contextualize them with local, sequence-level information to learn robust representations for scalable analysis. The remainder of this section is presented as follows. First, we introduce the necessary background knowledge and representations to understand the proposed approach. We follow with an overview on the different components in the framework and their contributions, before going deeper into each of the components. 

\subsection{Background: Metagenomes and Representations}
First, we introduce some necessary background to understand the \textit{scalable} metagenome analysis problem. A metagenome sample consists of a collection of $r$ nucleotide sequence \textit{reads} $\mathcal{X}$, where each \textit{read} $X_i \in \mathcal{X}_r$ belong to an organism $y_i\in \mathcal{Y}_m$. A metagenome sequence read $X_i = \{x_0, x_1, \ldots x_N\}$ is a sequence of $N$ nucleotide bases drawn from a set $\{A, T, C, G\}$. 
The \textit{scalable} metagenome analysis problem aims to learn a robust $D$-dimensional representation $h_{i}\in\mathbb{R}^{1\times D}$ for a given metagenome sequence read $X_i \in \mathcal{X}_r$ to identify its originating organism $y_i\in \mathcal{Y}_m$, with limited annotations. 
In this work, we consider two problem settings to evaluate the scalability and generalization capacity of metagenome analysis frameworks: (i) \textit{targeted} and (ii) \textit{generalized} metagenome analysis. 
In the \textit{targeted} setting, the evaluation is done in a \textit{supervised} learning setting where a limited set of \textit{labeled} \textit{clinical} metagenome samples are available for training. The clinical samples are labeled sequences with a set of known, closely related pathogens that follow a highly unbalanced distribution. 
In the \textit{generalized} setting, the problem is framed as an unsupervised learning problem, where novel, previously \textit{unseen} pathogens are introduced into the \textit{evaluation data}. 
The former setting evaluates the effectiveness of the learned representations for learning with limited \textit{labeled} data while the latter tests the capability to isolate \textit{novel} pathogen sequences to enable discovery and learning of new classes, \textit{with limited human supervision}. 
Combined, these two evaluation settings allow for robust evaluation of machine learning-based metagenome analysis frameworks in real-world settings. 

\subsection{Overall Approach}\label{sec:overallApproach}
We propose \textit{MG2Vec}, a self-supervised representation model that captures both structural and contextual properties from unlabeled metagenome sequences. 
The overall approach is illustrated in Figure~\ref{fig:arch}. Our approach consists of three main components: (i) the construction of a weighted, directed graph from $\mathcal{X}_r$ sequence reads to capture any structural dependencies within the nucleotide co-occurrences, (ii) learning an intermediate representation to capture the global structural priors from the constructed graph structure, and (iii) use an attention-based formalism to \textit{contextualize} and enhance the global structural representations, with local, sequence-level context. 
As can be seen from Figure~\ref{fig:arch}, the first step is to construct a graph from the metagenome sequences that provide a global structural prior for learning robust representations. 
We draw inspiration from the success of De Bruijn graphs~\cite{lin2016assembly} for genome assembly and construct a \textit{weighted, directed} graph from the metagenome sequences. 
Each node in the graph represents a k-mer~\cite{marccais2018asymptotically}, a subsequence from a genome read of length $k$. Note that this is different from the traditional De Bruijn graphs, which are also directed, but \textit{unweighted} multigraph and contain additional edges linking to auxiliary nodes representing ($k{-}1$)-mers and ($k{+}1$)-mers. The direction of the edge is given by the direction of the co-occurrence of k-mers in a given sequence read. The weight of each edge is a function of the frequency existence of an edge between two k-mers detected in metagenome sequences $\mathcal{X}$. Hence, the strength of each edge reflects the \textit{global} frequency of detection of the connection between two k-mers. The construction of the graph is a \textit{streaming} process, with nodes and edge weights updated on the observation of a genome sequence read. Inspired by ConceptNet~\cite{liu2004conceptnet}, the weights between nodes $x_i$ and $x_j$ are updated based on the function given by 
\begin{equation}
    w_{ij} = \lambda_{max} \sqrt{max(\Psi_{i,j} - 1, 1)} + d(\Psi_{i,j}, \lambda_{min})
    \label{eqn:weight_update}
\end{equation}
where $\Psi(x_i, x_j) {=} \frac{e_{i,j}}{(\Vert e_{i,j} - e^{\prime}_{i,j} \Vert_2)}$ and captures the relative increase in the co-occurrence frequency between the k-mers $x_i$ and $x_j$; $e_{i,j}$ is the current edge weight between $x_i$ and $x_j$, and $e\prime_{i,j}$ is the new weight to be updated; $d(\cdot)$ is a function that negates updates in the edge weights for assertions that are weakly reinforced through rare occurrence and is given by $d(\cdot) {=} min(\Psi_{i,j} {-} \lambda_{min}, \lambda_{min}) {+} \lambda_{min}$; and $\lambda_{min}$ and $\lambda_{max}$ are hyperparameters to control normalization bounds. In our experiments, we set both to $2$. 
Algorithm~\ref{alg:1} illustrates the pseudo-code. The algorithm takes as input: A metagenome sample $\mathcal{X}$, which consists of a collection of $r$ nucleotide sequence \textit{reads} $X_i \in \mathcal{X}_r$ belong to an organism $y_i\in \mathcal{Y}_m$. 
A metagenome sequence read $X_i = \{x_0, x_1, \ldots x_N\}$ is a sequence of $N$ nucleotide base pairs drawn from a set $\{A, T, C, G\}$. 
Other inputs include: the $k$ for constructing the k-mer nodes, the hyperparameters $\lambda_{min}$ and $\lambda_{max}$. 
The output is a structural graph that captures the global structure of the metagenome sample. Note that the unobserved k-mers will not be part of the graph structures. Hence each node will have at least one edge and no isolated nodes will be in the final graph. Also note that this is not a De Bruijn graph which will have additional nodes indicating  $k-1$-mers.

\begin{algorithm}[t]
\textbf{Input:} Metagenome Sequence sequence reads $X_i = \{x_0, x_1, \ldots x_N\}, \forall X_i \in \mathcal{X}_r$, $k$, $\lambda_{max}$, $\lambda_{min}$

\textbf{Output:} Global structural graph $G_S(V, E)$

\begin{algorithmic}[1]
\Procedure{getWeight}{$e_{i,j}, e^{\prime}_{i,j},  \lambda_{max},  \lambda_{min}$}

 \State $\Psi(x_i, x_j) {=} \frac{e_{i,j}}{(\Vert e_{i,j} - e^{\prime}_{i,j} \Vert_2)}$
 
 \State $w_{ij} = \lambda_{max} \sqrt{max(\Psi_{i,j} - 1, 1)} + d(\Psi_{i,j}, \lambda_{min})$
 \State $e_{i,j} = w_{ij}$ 
 \State \textbf{return} $e_{i,j}$
\EndProcedure

\State $G \gets \emptyset $ \Comment{Order-zero (null) Graph}
\State $\lambda_1 = \lambda_{max}$
\State $\lambda_2 = \lambda_{min}$
\For{$ X_i \in \mathcal{X}_r $} \Comment{Iterate through each read}
    \For{$ x_j \in X_i $} \Comment{Iterate through each nucleotide}
        \State $v_{m} \gets x_{j:j+k}$ \Comment{K-mer of length $k$ with stride 1}
        \If{$G ==  \emptyset$}
            \State $G \cup v_m$
        \Else
            \If{$v_{m} \not\in G$}
            \State $G \cup (v_{m-1}, v_m, 1)$
            \Else
            \State $w_{ij} {=} \textsc{getWeight}(e_{{m-1}, m}, 1, \lambda_{1},  \lambda_{2})$
            \State $G(v_{m-1}, v_m) \gets (v_{m-1}, v_m, w_{ij})$ 
            \Comment{Update weight of edge connecting $v_{m-1}, v_m$}
            
            \EndIf
        \EndIf
    \EndFor
\EndFor
\State \textbf{return} $G$
\end{algorithmic}
\caption{Global Structural graph construction. A metagenome sample $\mathcal{X}$, which consists of a collection of $r$ nucleotide sequence \textit{reads} $X_i \in \mathcal{X}_r$ belong to an organism $y_i\in \mathcal{Y}_m$. A metagenome sequence read $X_i = \{x_0, x_1, \ldots x_N\}$ is a sequence of $N$ nucleotide base pairs. The output is a structural graph that captures the global structure of the metagenome sample.}\label{alg:1}
\end{algorithm}

\begin{figure*}
    \centering
    \begin{tabular}{ccc}
         \includegraphics[width=0.29\textwidth]{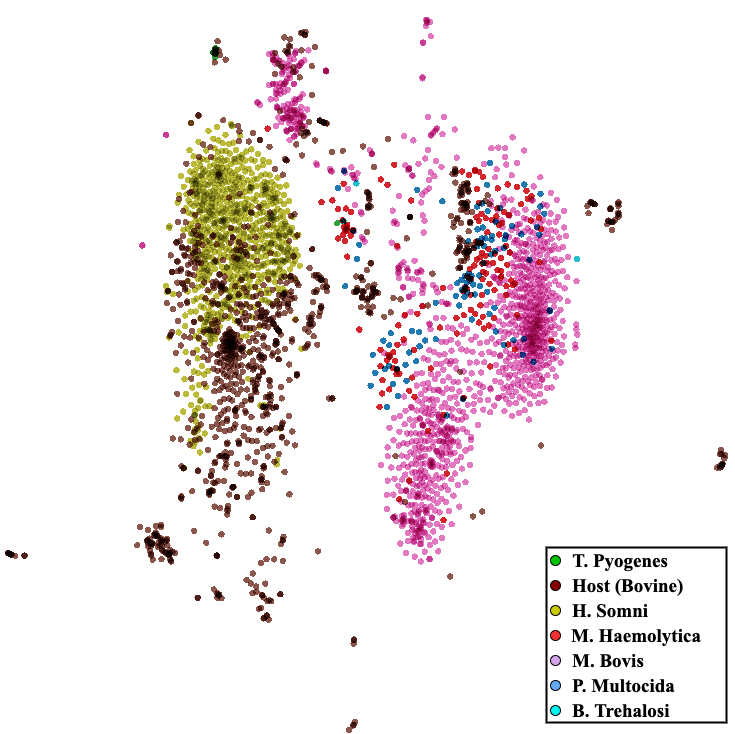} & 
        \includegraphics[width=0.29\textwidth]{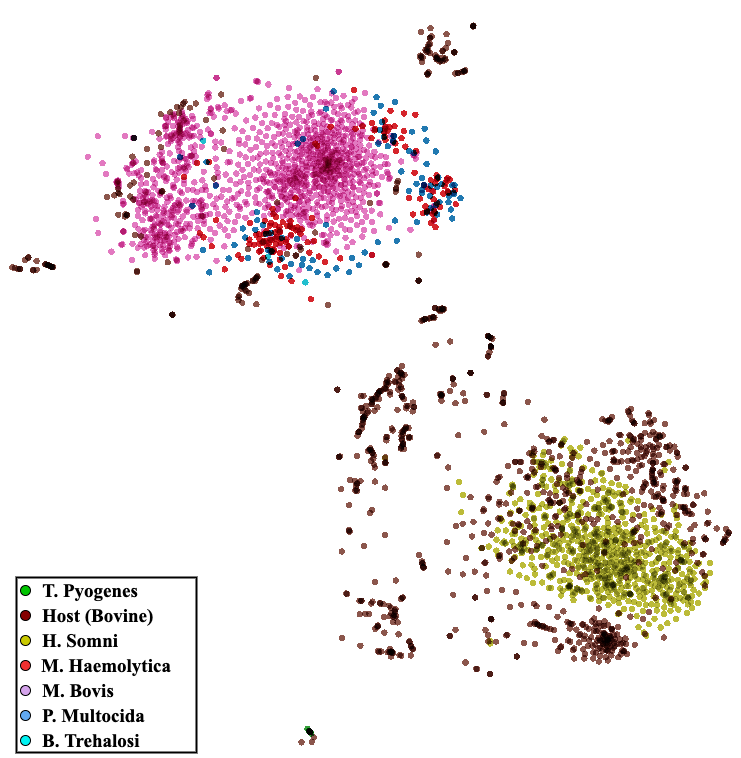}
         \includegraphics[width=0.29\textwidth]{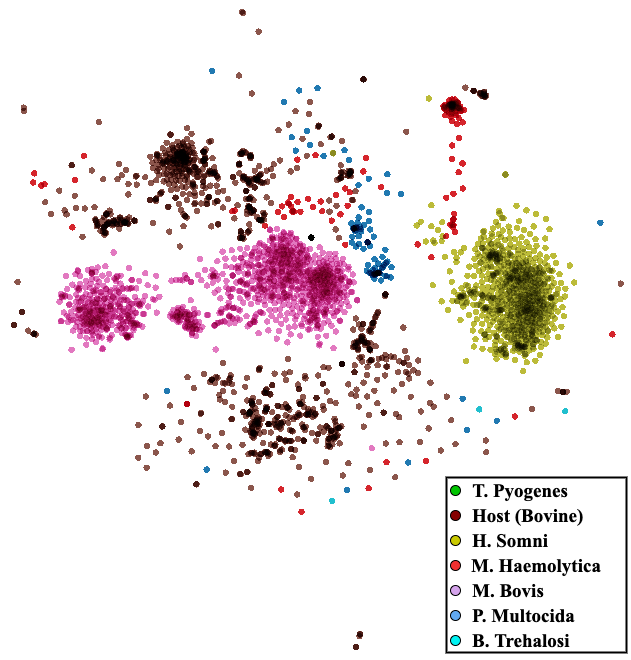}\\
         (a) & (b) & (c)\\
    \end{tabular}
    \caption{\textbf{Visualization} of the features from (a) global structural priors, (b) contextualized sequential representations and (c) \textit{MG2Vec} representations that explicitly code structural and contextual representations. \textit{MG2Vec} provides a much cleaner segmentation between classes compared to just structural or just contextual features.}
    \label{fig:tSNE}
\end{figure*}

\subsection{Global Structural Priors}\label{sec:globalStructure}
Learning a low-dimensional representation of the global structural prior in a metagenome is the next step in our proposed framework. This representation must ideally have the following properties: (i) capture the structural similarity between nodes (k-mers) such that they share a set of ``roles'' within the genome sequence and (ii) capture the ``community'' or neighborhood structure of a k-mer to enable the retrieval of relevant nodes due to any noisy sequence reads, which is not that uncommon in next-generation sequencing approaches~\cite{laver2015assessing}. To construct global structural representations with these properties, we use Node2Vec~\cite{grover2016node2vec} as a mechanism to capture the global structure of the metagenome sequences. The global structural representations are constructed using the Skip-Gram approach~\cite{mikolov2013distributed} to capture network structure by the following objective function to maximize the log-probability of observing a structural k-mer neighborhood $N_S(x_i)$ for a node $x_i$ conditioned on its feature representation, given by $h^g_i$ as follows:
\begin{equation}
    \argmax_{h^g_i} \sum_{x_i\in G_r} log \: Pr(N_S(x_i) | h^g_i)
    \label{eqn:globalObjective}
\end{equation}
where $G_r$ is the weighted, directed graph constructed from metagenome sequences $\mathcal{X}$ and $x_i$ is a node corresponding to a unique k-mer. Note that since we construct the graph for the entire metagenome and the edge weights are aggregated using the function from Equation~\ref{eqn:weight_update} and this optimization is done over each node \textit{within} each sequence and averaged across the entire metagenome. The neighborhood $N_S$ is generated by a biased random walk whose transition probabilities are a function of the weight between k-mers (nodes). Hence the neighborhood of each node (k-mer) is a possible pathogen-specific genome sequence that can be found within an overall metagenome infected by the pathogen. The normalization of the edge weights (from Equation~\ref{eqn:weight_update}) allows us to balance the transition probabilities across multiple walks and hence develop a more robust structural prior for each k-mer. 
We empirically verify with experimental results in Section~\ref{sec:ablation} to corroborate this assertion. 

\subsection{Contextualized Representations with Attention}\label{sec:pre_training}
The next step in building robust metagenome representations is \textit{contextualization}. We consider a representation to be \textit{contextualized} if the resulting representation can (i) reduce the impact of background or spurious structural or sequential patterns that may occur due to noisy observation and (ii) \textit{integrate} a given k-mer's feature representation with its surrounding context. This representation is analogous to a posterior-weighted representation that highlights areas of interest while suppressing contextually irrelevant features. 
To build such representations, we use the idea of \textit{word embedding} to learn a mapping function $M_c{:} x_i\rightarrow \mathbb{R}^{1\times D}$. However, to ensure that the context is captured effectively we leverage the attention-based transformer~\cite{vaswani2017attention} to learn our contextualized mapping function, which has become a standard backbone network for learning contextualized, semantic representations in many natural language processing tasks~\cite{devlin2019bert,radford2019language}. 
We initialize the mapping function $M$ with the global structural embedding from Section~\ref{sec:globalStructure} and use the transformer to capture contextual properties in each sequence. 

A transformer is a neural network architecture composed of a series of multi-headed self-attention (MHA) and token-wise feedforward operations. At each level of the network, the representation of each token is updated based on the hidden representation from the lower levels. In our architecture, the k-mers, extracted from each sequence read ($\mathcal{X}_j{=}x_1, x_2, \ldots x_N; \forall X_j {\in} \mathcal{X}_r$), are fed as tokens to the network and the hidden representations of the final layer $H{=}h_1, h_2, \ldots h_D$ represent the learned, contextual representations for each k-mer. The transformer representations are conditioned on the global structural prior by initializing the embedding layer with the mapping function $M$. 
To ensure that these representations further capture the contextual properties from the entire sequence, we allow the model to use bidirectional attention by employing a masking approach, as common in masked language models such as BERT~\cite{devlin2019bert}. Specifically, we randomly \textit{mask} or replace a subset of tokens in the input $X_j$ with a [MASK] token and train the network to predict these missing tokens. Hence the objective function for the network is to minimize the log-likelihood of the masked tokens $X\mathcal{X}^\Gamma_j$ given the observed tokens ($\mathcal{X}^{-\Gamma}_{j}$) and is represented as:
\begin{equation}
    \mathcal{L}(\mathcal{X}^\Gamma_j | \mathcal{X}^{-\Gamma}_{j}) = \frac{1}{T}\sum_{t=1}^{T}log \: p(x^{\Gamma}_{t} | \mathcal{X}^{-\Gamma}_{j};\theta)
    \label{eqn:mlm_objective}
\end{equation}
where each genome sequence is given by $\mathcal{X}_j{=}\mathcal{X}^{-\Gamma}_{j}\bigcup \mathcal{X}^{\Gamma}_{j}$ and $\mathcal{X}^{\Gamma}_{j}{=}\{x^{\Gamma}_1, \ldots x^{\Gamma}_t\}$ is a set of $T$ masked tokens; $\theta$ is the set of learnable parameters in the network. The masks are chosen such that the probability of masking is given by $p(M){=}s^T(1-s)^{N-T}$ where $s$ is the masking ratio. 

The objective function from Equation~\ref{eqn:mlm_objective} is used to train the transformer architecture, and more importantly, the errors are allowed to propagate all the way to the initial mapping function $M$, which is initialized with the global structural priors. The updated mapping function ($M_c$) provides allows us to generate \textit{contextualized} embeddings for each k-mer in a sequence and allow us to provide a balance between the global structural prior from the \textit{metagenome} and the local contextual features within each genome sequence read. The final \textit{MG2Vec} representation a sequence read $h^m_i$ is then constructed by the concatenation of the global representation $h^g_i$ and the contextual representation $h^c_i$ for a given k-mer $x_i$. Hence, $h^m_i{=}[h^g_i;h^c_i]$ where $h^g_i{=}M(x_i)$ and $h^c_i{=}M_c(x_i)$.
Note that this is different from the current use of transformers~\cite{devlin2019bert,radford2019language} where the output of the encoder $H$ is used as the extracted features. 
Our approach explicitly encodes the global structural, and local contextual properties of k-mers within a metagenome is similar in spirit to ELMo~\cite{peters2018deep}, which aims to capture the context-dependent aspects of word meaning using bidirectional LSTMs~\cite{hochreiter1997long}. 
To highlight the importance of such explicit representations, we visualize a subset of clinical metagenome sequences for all three representations - global, contextual and \textit{MG2Vec} using TSNE~\cite{van2008visualizing} in Figure~\ref{fig:tSNE}. As can be seen, the explicit representation of structural and contextual properties allows for cleaner segmentation between sequence reads across species. 
Experimental evaluation (presented in Section~\ref{sec:results}) corroborate the observation from visual inspection.

\subsection{Implementation Details}\label{sec:impl_details}
In our experiments, we use $k=4$ to extract k-mer representations with a stride of $1$ on sequence reads. We use this setting based on genome analysis studies that have shown that tetranucleotide bases from phylogenetically similar species can be very similar between closely related species~\cite{perry2010distinguishing}. 
The resulting structural graph has $1296$ nodes, which forms our vocabulary. 
We use Node2Vec on the global structural graph to obtain structural priors. The number of walks per source node is set to $10$, and the walk length is set to $80$. 
We use a transformer with 4 hidden layers and 8 attention heads with a hidden size of $512$ and a dropout rate of $0.1$. We pre-train for $10$ epochs on two unlabeled metagenome samples with a combined total of $2{,}151{,}436$ sequence reads and a batch size of $16$ and the learning rate schedule from~\cite{vaswani2017attention}. 
The training was done on a server with 2 Titan RTXs and a 64 core AMD CPU and took around $36$ hours to converge. 
\section{Experimental Setup}\label{sec:exp_setup}
In this section, we describe the evaluation setup to assess the performance of our framework in detail. Specifically, we outline the different evaluation settings that we propose to evaluate its pathogen detection performance, along with a discussion on the primary pathogens considered for targeted and generalized pathogen detection evaluation. We conclude with a discussion on the metrics and baselines used to benchmark the approach. 

\subsection{Evaluation settings} \label{sec:settings}
We evaluate our approach on four different settings with varying levels of difficulty. Specifically, we first assess the model's performance on targeted pathogen detection, where the goal is to learn to recognize \textit{known} pathogen sequences from curated metagenomes sequences under a constrained setting where distractor sequences are removed i.e., sequences that are not unique to a single pathogen are removed. Second, we assess the model's performance on \textit{unconstrained} targeted pathogen detection, where the goal is to identified known pathogen sequences from millions of sequences where the metagenome sequences are not curated i.e., the clinical samples are sequenced and presented directly to the algorithm. The amount of unique sequences to the target pathogen can be extremely small (sometimes in the order of $10^{-4}$). Third, we evaluate the generalization capacity of the representations by presenting the approach with metagenome sequences that has both unknown pathogens, but related pathogens, and known pathogens. Finally, we evaluate its generalization capability when presented with large-scale data that is completely out of domain. The former two experiments are designed to evaluate the capability of the approach to detect known pathogens from clinical samples when labeled training data is available. The latter two are designed to assess the generalization capability of the proposed model that can enable novel pathogen discovery with minimal supervision. 

\subsubsection{Pathogens used for evaluation}\label{sec:pathogen_list}
We consider the pathogens associated with the Bovine Respiratory Disease Complex (BRD) as our primary set of pathogens for targeted and generalized pathogen detection benchmarks. This complex multi-etiologic disease affects cattle worldwide and is one of the leading causes of economic distress in the cattle industry. 
The chief bacterial pathogens of interest are \textit{Mannheimia haemolytica}, \textit{Pasteurella multocida}, \textit{Bibersteinia trehalosi}, \textit{Histophilus somni}, \textit{Mycoplasma bovis}, and \textit{Trueperella pyogenes}, which mostly belong to the \textit{Pasteurellaceae} family. 
For generalized pathogen evaluation (Section~\ref{sec:generalDetection}), we considered a simulated metagenome with the following \textit{viral} and bacterial pathogens: Bovine viral diarrhea virus (BVDV), Bovine parainfluenza virus 3 (BPIV-3), Bovine herpesvirus 1 (BoHV-1), Bovine coronavirus (BCoV) and Bovine respiratory syncytial virus (BRSV). 
In addition to the pathogens from the Bovine Respiratory Disease Complex (BRDC), to evaluate for true out-of-domain generalization, we also consider a separate dataset~\cite{liang2020deepmicrobes} containing $2505$ pathogens found in the human gut microbiome from the ENA study accession ERP108418~\cite{almeida2019new}. 

\begin{table*}[ht]
    \centering
    \caption{\textbf{Targeted, constrained pathogen detection.} Performance of different machine learning baselines on the recognition task on curated, unbalanced metagenome data using the MG2Vec representations. Precision (\textit{Prec.}) and Recall (\textit{Rec.}) are reported for each class.}
    \resizebox{0.99\textwidth}{!}{
    \begin{tabular}{|c|c|c|c|c|c|c|c|c|c|c|c|c|c|c|}
    \toprule
    \multirow{2}{*}{Approach} & \multicolumn{2}{|c|}{\textbf{Host}} & \multicolumn{2}{|c|}{\textbf{H. somni}} & \multicolumn{2}{|c|}{\textbf{M. bovis}} & \multicolumn{2}{|c|}{\textbf{M. haemolytica}} & \multicolumn{2}{|c|}{\textbf{T. pyogenes}} & \multicolumn{2}{|c|}{\textbf{P. multocida}} & \multicolumn{2}{|c|}{\textbf{B. trehalosi}} \\
    \cline{2-15}
     & Prec. & Rec. & Prec. & Rec. & Prec. & Rec. & Prec. & Rec. & Prec. & Rec. & Prec. & Rec. & Prec. & Rec.\\
    \toprule
    LR & 0.97 & \textbf{0.99} & 0.80 & 0.68 & 0.86 & 0.94 & 0.65 & 0.61 & \textbf{0.85} & 0.73 & 0.00 & 0.00 & 0.00 & 0.00 \\ 
    SVM & 0.97 & 0.98 & 0.75 & 0.66 & 0.82 & \textbf{0.96} & \textbf{0.67} & 0.34 & 0.67 & 0.13 & 0.00 & 0.00 & 0.00 & 0.00\\ 
    MLP & 0.98 & \textbf{0.99} & 0.82 & \textbf{0.77} & \textbf{0.89} & 0.94 & 0.60 & 0.69 & 0.78 & 0.93 & 0.00 & 0.00 & 0.00  & 0.00 \\ 
    DL  & \textbf{0.99} & 0.97 & \textbf{0.84} & 0.73 & 0.87 & 0.95 & 0.44 & \textbf{0.82} & 0.68 & \textbf{1.00} & \textbf{0.50} & \textbf{0.26} & \textbf{ 0.10} & \textbf{0.33} \\ 
    \midrule
    K-Means & 0.95 & 0.36 & 0.33 & 0.75 & 0.55 & 0.87 & 0.24 & 0.12 & 0.62 & 1.00 & 0.47 & 0.41 & 0.00 & 0.00 \\ 
    \bottomrule
    \end{tabular}
    }
    \label{tab:quant_result}
\end{table*}

\subsection{Metrics and Baselines}
To quantitatively evaluate our approach, we use precision, recall, and the F1-score for each class. We do not use accuracy as a metric since real-life metagenomes can be highly skewed towards host sequences, and it would be possible to obtain high accuracy ($>90\%$) by only predicting the dominant class (i.e., the host). Precision and recall, on the other hand, allow us to quantify the false alarm rates and provide more precise detection accuracy. 
We compare against comparable representation learning frameworks for metagenome analysis proposed in literature such as graph-based approaches~\cite{narayanan2020genome}, and a deep learning model termed S2V, an end-to-end sequence-based learning approach based on DeePAC~\cite{bartoszewicz2020deepac}. The S2V baseline is an adaptation of the models used in DeePAC~\cite{bartoszewicz2020deepac} and DeepMicrobes~\cite{liang2020deepmicrobes} for the single-read shotgun sequencing problem setup. For classification, we consider a balanced mix of both traditional machine learning baselines and deep learning baselines. Traditional baselines include Support Vector Machines (SVM), Logistic Regression (LR), and a feed-forward neural network (NN) with two hidden layers with 256 neurons each. We consider two deep learning baselines. The first is a deep, feed-forward neural network with 3 hidden layers with 256, 512 and 1024 neurons each with a ReLU activation function. A residual connection is added between layers 1 and 3. The second deep learning baseline is a 1-layer LSTM model with 64 neurons in the hiddden layer and a recurrent dropout layer with a dropout probability of 0.4. A dense layer with 32 layer is used to project it onto a learned space before classification. The k-mers are represented using a word2vec model~\cite{ling2015two} trained on the data. 
We choose the hyperparameters for each of the baselines using an automated grid search and the best performing models from the validation set were taken for evaluation on the test set. 
The final hyperparameters for each baseline are as follows:
\begin{enumerate}
    \item \textit{Logistic Regression}: This baseline was trained using the Newton-CG solver with an L2 penalty and the maximum number of iterations was set to 10000. The regularization was set to 0.1. 
    \item \textit{Support Vector Machine}: The SVM baseline was trained in a one-versus-one setting for the multi-class setting. A polynomial kernel was used with a maximum interation setting of 10000. A squared L2 penalty was used with a regularization of 0.1.
    \item \textit{MLP}: A 2-layer neural network with 256 neurons each with ReLU activation was trained with the Adam optimizer and an initial learning rate of $4\times 10^{-5}$ and a L2 regularization with penalty of $1\times 10^{-5}$.
    \item \textit{Deep Learning}: An Adam optimizer was used with a learning rate scheduled given by the custom training schedule in the original transformer paper~\cite{vaswani2017attention}. Due to the unbalanced nature, a weighted cros-entropy loss was used with the class weights set as the normalized inverse of the percentage of examples for each class. It was trained for 10 epochs with a batch size of 64 and for 15 epochs with a batch size of 512. 
\end{enumerate}

We use the Sci-Kit Learn package~\cite{pedregosa2011scikit} for the traditional machine learning classification models and Keras~\cite{chollet2015keras} for the deep learning baselines.

\section{Experimental Evaluation}\label{sec:results}
In this section, we present the experimental evaluation results for the approach on the four settings outlined in Section~\ref{sec:settings}. In each setup, we describe the data collection process, the evaluation setup and any additional baselines.

\subsection{Targeted Pathogen Detection}
We first present results under the \textit{targeted} pathogen detection setting, where the goal is to recognize metagenome sequences belonging to pathogens that were known during both pre-training and finetuning stages.

\begin{figure*}
    \centering
    \begin{tabular}{ccc}
    \includegraphics[width=0.31\textwidth]{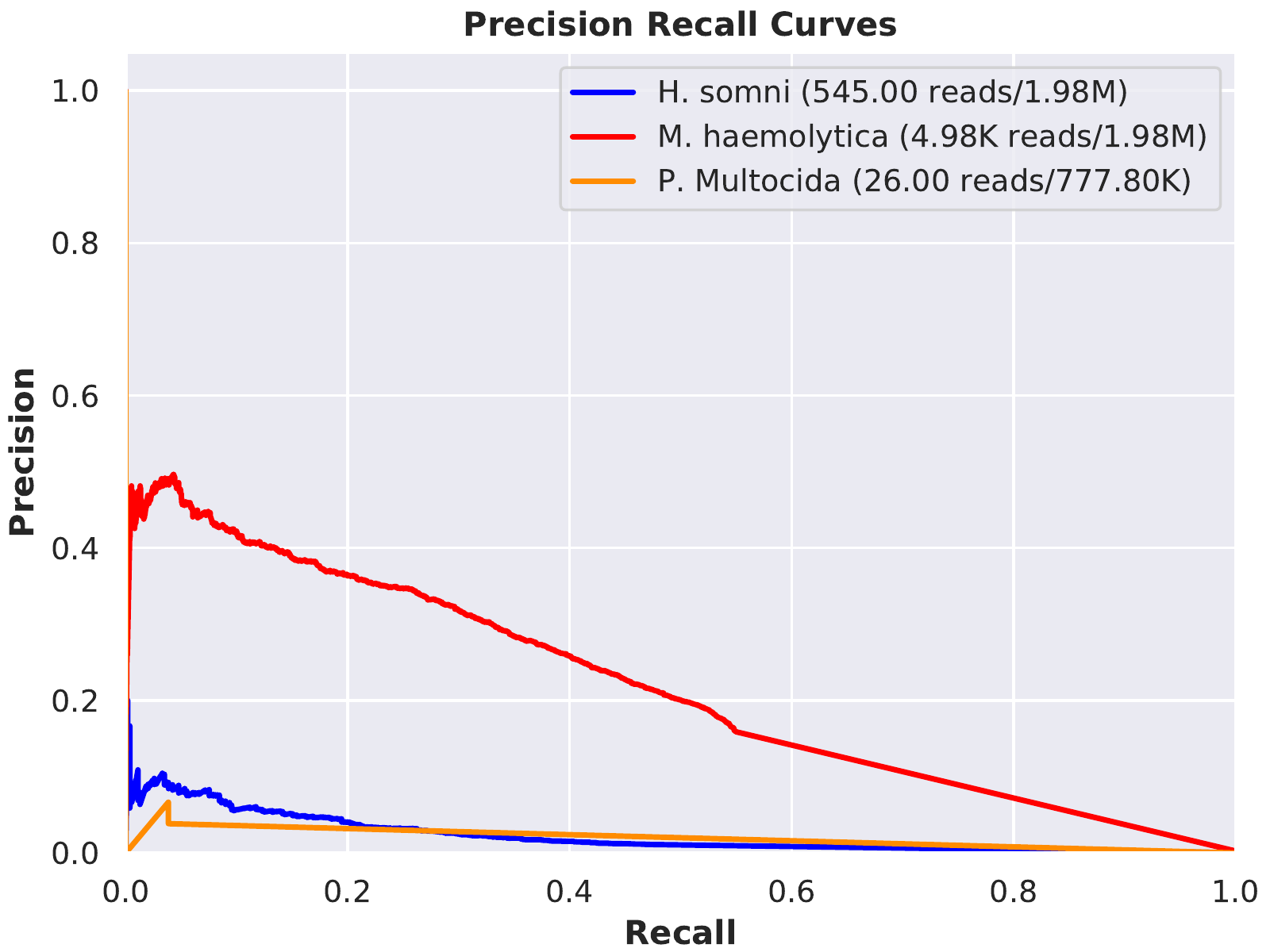} & 
    \includegraphics[width=0.31\textwidth]{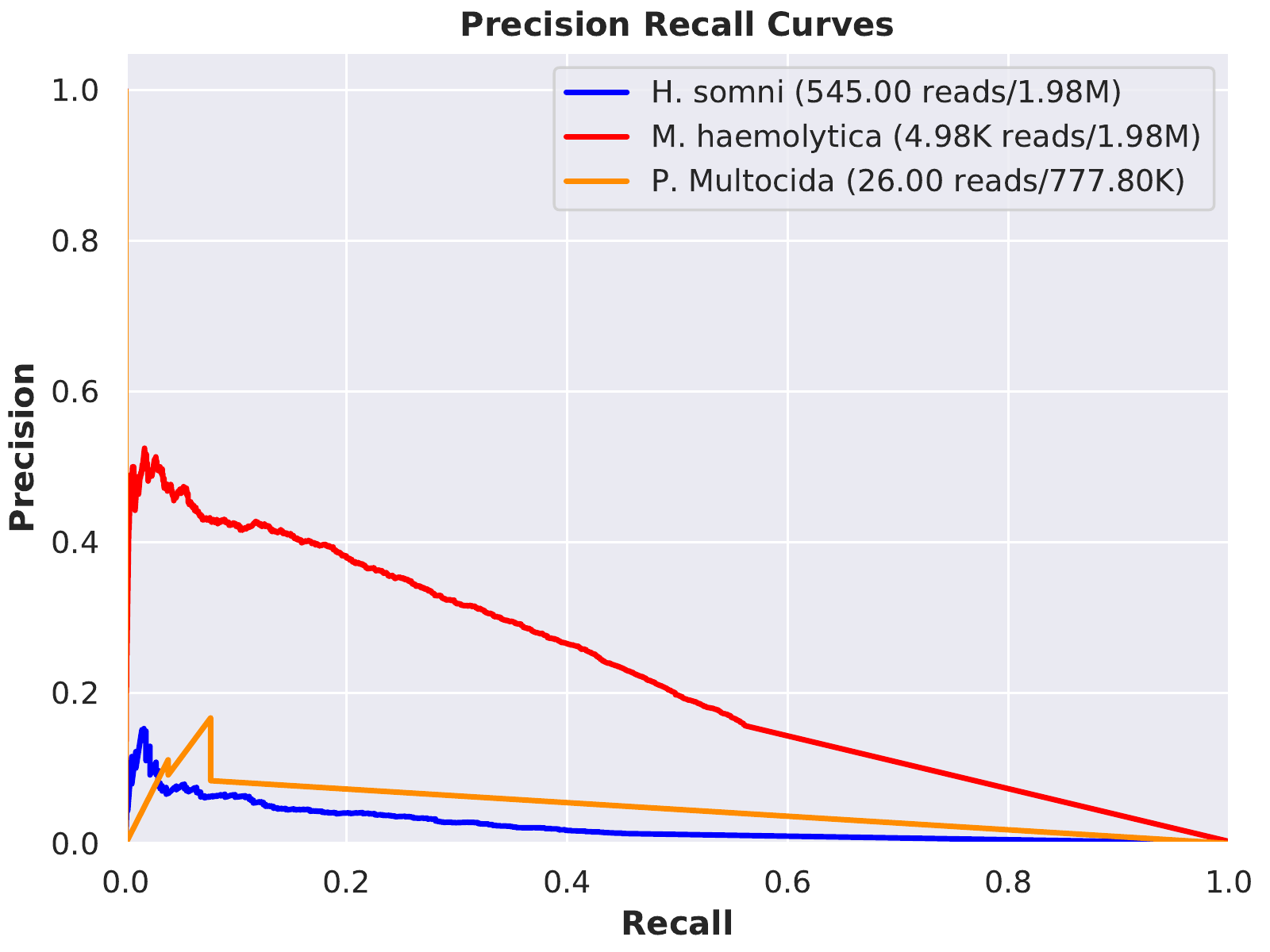} & 
    \includegraphics[width=0.31\textwidth]{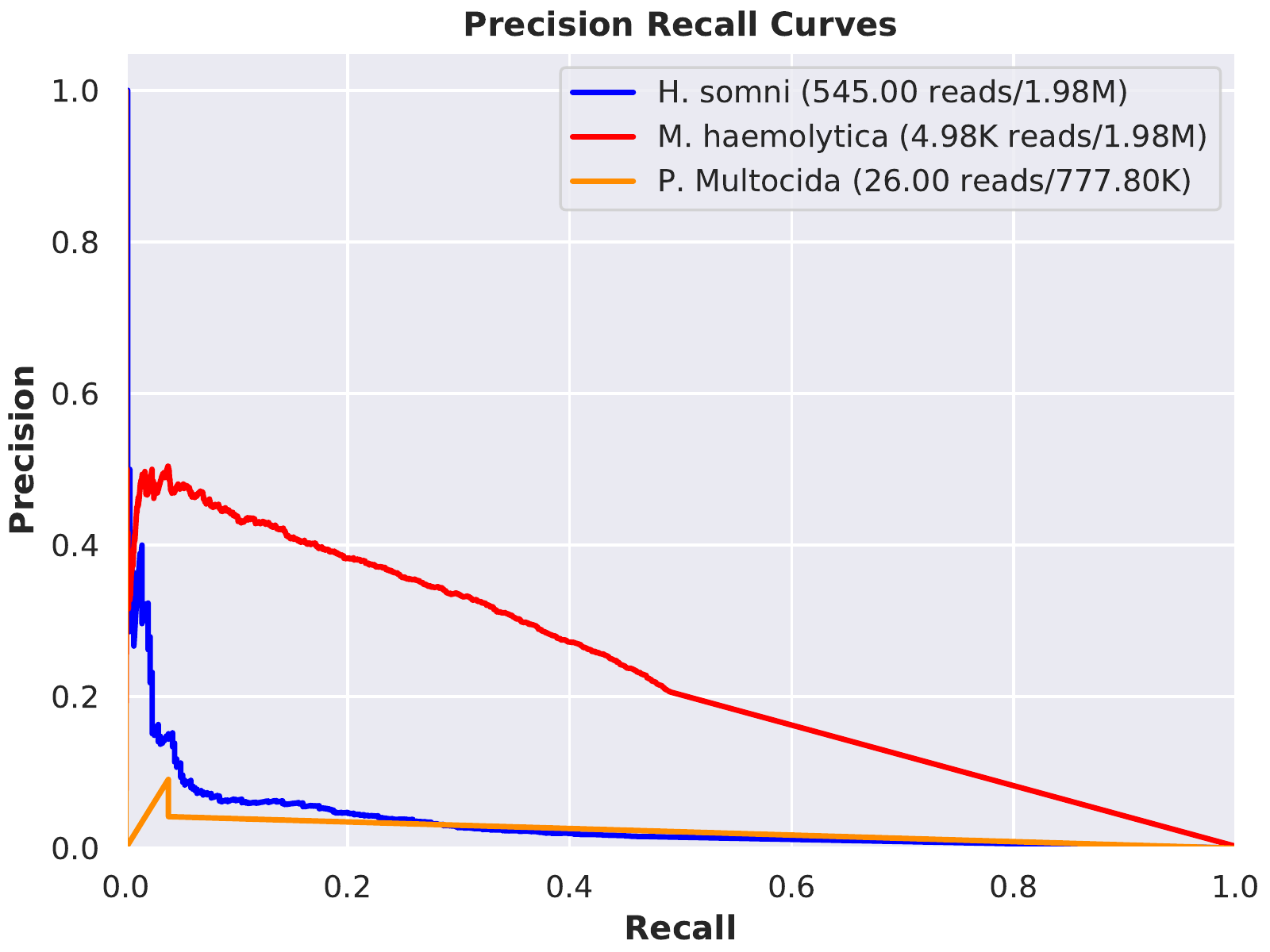} \\
    (a) & (b) & (c)\\
    \end{tabular}
    \caption{\textbf{Unconstrained known pathogen detection.} The precision-recall curves of (a) GN-2V, (b) C2V and (c) MG2Vec are shown. All results are shown on entire, uncurated metagenomes that were independently tested with PCR and culture tests for pathogen presence.}
    \label{fig:unconstrained_prc}
\end{figure*}

\subsubsection{Constrained Evaluation}\label{sec:exp1}
In this evaluation setting, we extract contextualized representations from real-world, clinical metagenome sequences to evaluate the performance of the proposed models and baseline models for targeted pathogen detection under constrained settings. We use the term ``\textit{constrained}'' to refer to the fact that we filter out sequences that occur in the genome of more than one species. Hence the resulting sequences are known to belong to exactly one of the host or bacterial pathogen species associated with BRDC.

\textbf{Data.} To obtain real-world metagenome data, we extracted sequences from $13$ Bovine Respiratory Disease Complex (BRDC) lung specimens at the Oklahoma Animal Disease Diagnostic Laboratory at Oklahoma State University. DNA sequences are extracted from lung samples using the DNeasy Blood and Tissue Kit (Qiagen, Hilden, Germany). Sequencing libraries are prepared from the extracted DNA using the Ligation Sequencing Kit and the Rapid Barcoding Kit. Prepared libraries are sequenced using MinION (R9.4 Flow cells), and sequences with an average Q-score of more than $7$ are used in the final genome.
Further quality assessment was done using RScript MinIONQC~\cite{lanfear2019minionqc}. Pathogens in the metagenome sequence data are identified using a modified version of the bioinformatics pipeline from Stobbe \textit{et al.}~\cite{stobbe2013probe}. First, unique pathogen fingerprints are extracted from each pathogen genome using the modified Tools for Oligonucleotide Fingerprint Identification (TOFI)~\cite{satya2008high}. 
Next, the unique pathogen fingerprints are aligned with the metagenome sequence data using a BLOSUM matrix~\cite{henikoff1992henikoff}. 
Sequences highly similar to the pathogen fingerprint (i.e., alignment score below $10^{-9}$) were labeled as specific pathogen sequences, and others were labeled as host sequences. Since this could label other pathogens as host, we randomly sample sequences belonging only to the host genome for precise evaluation. 
Samples from $7$ patients were used for training, $5$ were used for evaluation, and the remaining for validation. 

The results are summarized in Table~\ref{tab:quant_result}. It can be seen that our \textit{MG2Vec} representations are robust enough to distinguish between highly similar pathogen sequences with smaller and simpler machine learning baselines such as logistic regression. In fact, a simple k-means clustering approach was able to achieve reasonable performance on this benchmark, indicating that the learned representations are relatively segmented in the feature space. Of particular interest is that there is a perfect recall on \textit{T. Pyogenes} which is one of two pathogens (the other being \textit{M. bovis}) not in the same family as the others. 
A deep learning model, trained with weighted cross-entropy, has the better overall performance across all pathogen sequences. With reasonable precision, it can identify \textit{P. Multocida} and \textit{B. Trehalosi}, which are the least represented classes in the training set with $37$ and $17$ labeled sequences, respectively. 

\begin{table}[t]
    \centering
    \caption{
    \textbf{Comparison with other representations.} Performance evaluation of machine learning baselines using other metagenome representations. Average F1 scores are reported.
    }
    \resizebox{0.99\columnwidth}{!}{
    \begin{tabular}{|c|c|c|c|c|c|c|c|c|}
    \toprule
    \multirow{2}{*}{\textbf{Approach}} & \multicolumn{2}{|c|}{\textbf{Node2Vec}} & \multicolumn{2}{|c|}{\textbf{SPK}} & \multicolumn{2}{|c|}{\textbf{S2V}} & \multicolumn{2}{|c|}{\textbf{MG2V}}\\
    \cline{2-9}
    & \textbf{Host} & \textbf{Path.} & \textbf{Host} & \textbf{Path.} & \textbf{Host} & \textbf{Path.}  & \textbf{Host} & \textbf{Path.} \\
    \toprule
    LR       & 0.824  & 0.044  & 0.857  & 0.128  & - & - & 0.979  & 0.509 \\
    SVM      & 0.806  & 0.071 &  0.863  & 0.113  & - & - & 0.975  & 0.405 \\
    MLP      & 0.849  & 0.097  & 0.864  & 0.080  & -  & - & \textbf{0.985}  & 0.534 \\
    DL       & 0.744  & 0.102  & 0.783  & 0.099  & 0.758 & 0.362 & 0.981  & \textbf{0.631}\\
    \bottomrule
    \end{tabular}
    }
    \label{tab:graph_rep_comp}
\end{table}

However, we find that these simple baselines do not always work in this task, as indicated by their performance when using other contemporary feature representations as seen from Table~\ref{tab:graph_rep_comp}. We compare the performance of the same baselines when presented with features (Node2Vec and Shortest Path Kernel~\cite{borgwardt2005shortest}) from GRaDL~\cite{narayanan2020genome}, an approach similar in spirit to ours, which proposes to use graph representations from \textit{individual} genome sequences for pathogen detection. We also compare against the sequence-based feature representation approach S2V, adapted from DeePAC~\cite{bartoszewicz2020deepac} (denoted as S2V) on our data. It can be seen that although the genome sequence representations are learned in a similar manner, they do not scale to the long-tail distribution in the limited, unbalanced labeled data.

\begin{figure*}
    \centering
    \begin{tabular}{ccc}
    \includegraphics[width=0.3\textwidth]{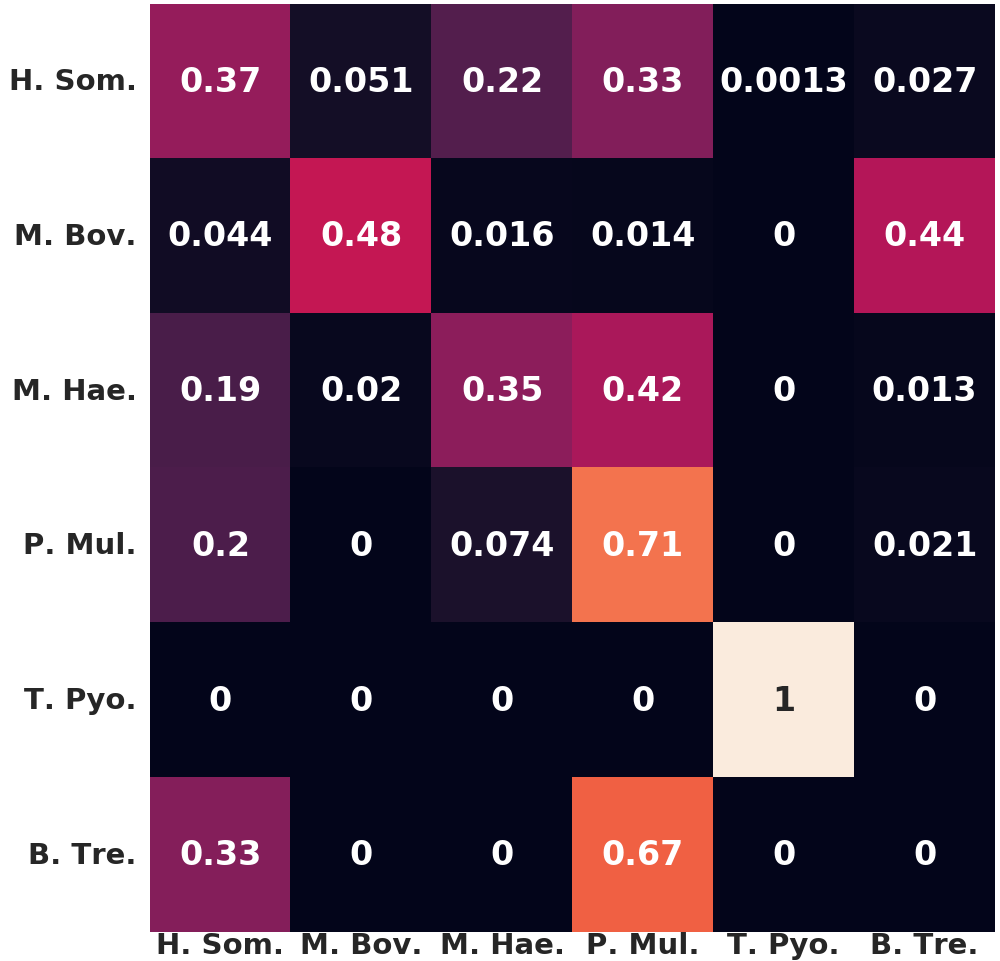} & 
    \includegraphics[width=0.305\textwidth]{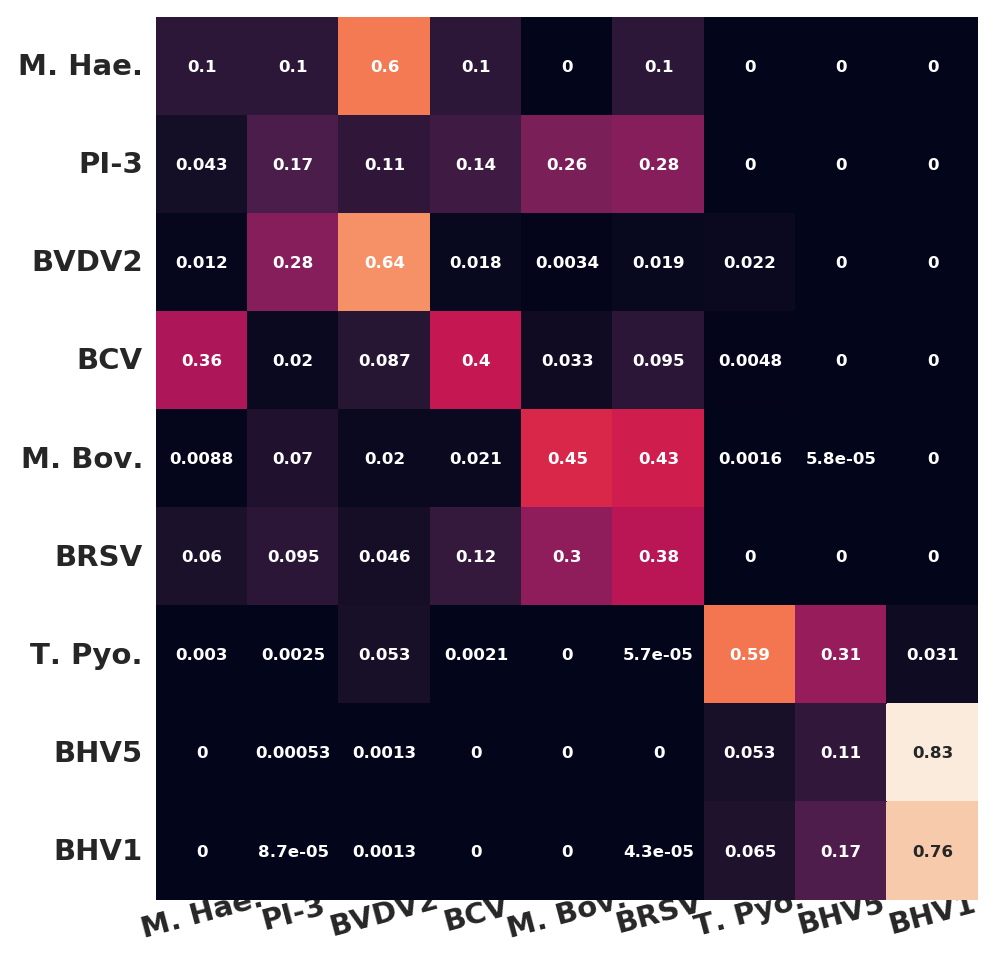} & 
    \includegraphics[width=0.302\textwidth]{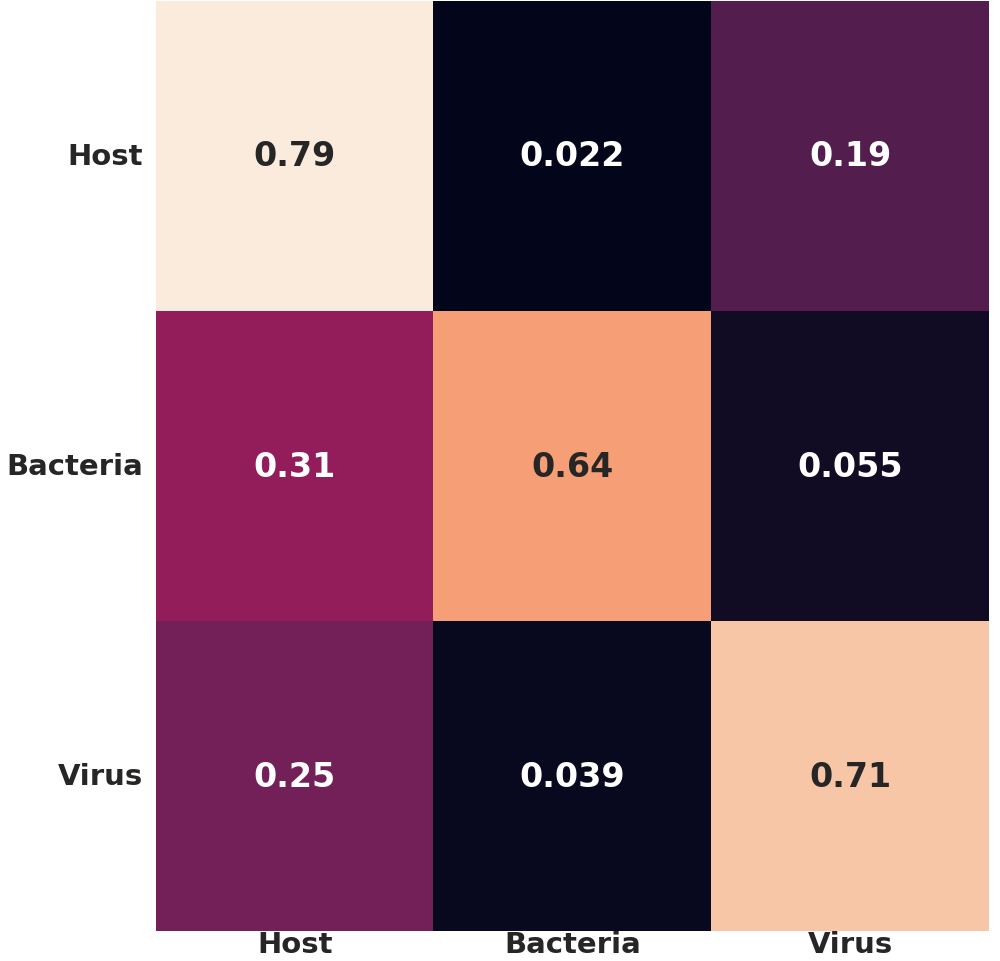} \\
    (a) & (b) & (c)\\
    \end{tabular}
    \caption{\textbf{Generalized Pathogen Detection.} Confusion matrices from unsupervised segmentation of a metagenome containing (a) only seen pathogens and (b) both seen and unseen pathogens, including unseen viruses. (c) shows the confusion matrix for unsupervised segmentation of a metagenome into pathogen groups. Note: results are showed on ``in-the-wild'' metagenomes with more than a million sequence reads.}
    \label{fig:pathogen_clustering}
\end{figure*}

\subsubsection{Unconstrained Evaluation}
To evaluate the performance on unconstrained pathogen detection, we present entire, uncurated metagenome samples to the deep learning model to detect the presence of three target pathogens - \textit{M. haemolytica}, \textit{H. somni}, and \textit{P. multocida}. The challenges presented in this evaluation are three-fold. First, since the metagenomes are uncurated, the distractor sequences (described in Section~\ref{sec:settings}) are not removed and hence will provide a significant challenge in identifying the target pathogens from the remaining sequences. Second, the sequence reads belonging to the host (Bovine) genome, will be significantly in number compared to the pathogen sequences, occupying more than $99\%$ of the sequence reads. Hence, the amount of false positives can potentially be very high when considering the detection rates of the three known pathogens. Finally, the three pathogens themselves are closely related, as seen from Table~\ref{tab:genome_sim}. Hence their genome can possess significant overlap which makes distinguishing between them more challenging. 

\textbf{Data.} To avoid any potential overlap with the training data (Section~\ref{sec:exp1}), we sequenced an independent set of 5 positive bovine lung samples submitted to the Oklahoma Animal Disease Diagnostic Laboratory (OADDL) for diagnostic assessment. Confirmation of the presence of the target pathogens was obtained through both traditional culture-based and PCR-based approaches. Culture strategies involve the use of different media such as Chocolate agar, Blood agar, MacConkey agar and PEA agar (Hardy Diagnostics, CA, USA). Isolated colonies were identified using the MALDI-ToF MS (Matrix Assisted Laser Desorption Ionization Time of Flight Mass Spectrometry). PCR confirmation was obtained using primers published previously by Kishimoto \textit{et al}~\cite{kishimoto2017development}. Note that these samples were chosen from a larger pool of specimen based on their Ct values, after calibration by an independent dilution study per standard procedure~\cite{gibson1996novel}. 
DNA was extracted from these samples using Qiagen DNeasy Blood and Tissue kit (Qiagen, Hilden, Germany). Sequencing libraries were made using library prep kits by Oxford Nanopore Technologies (ONT) viz., ligation sequencing kit (SQK-LSK109, ONT, UK), Rapid Barcoding kit (SQK-RAB004, ONT, UK) and field sequencing kit (SQK-LRK001, ONT, UK). 

\textbf{Evaluating Correctness.} We use BLAST[CITE], often considered as a gold-standard tool to identify and classify sequences~\cite{mcintyre2017comprehensive}, to evaluate the correctness of the predictions made by the baselines approaches. BLAST uses BLOSUM (BLOcks Substitution Matrix)~\cite{heinzinger2019modeling} or the PAM (Point Accepted Mutation)~\cite{dayhoff1972model} matrix to assess similarities between the query sequence and the database. From the matrix, BLAST draws initial raw scores, based on the extent of similarity between query and database. To facilitate alignments, BLAST introduce gaps and thus to assess homology between query and database that includes gaps and probable mutations, ‘bit scores’ are introduced[24]. In this experiment, we use the database with only the \textit{M.haemolytica genome} and thus bit scores can be accurate indicators for query matches to the database. The \textit{e-value}, computed from the bit score, act as an indicator about the significance of the match between nucleotides in query and database~\cite{NCBI,kerfeld2011using}. We consider e-values of 0.01 or lower to be a good measure of a match. 

\textbf{Quantitative Results.} We evaluate the proposed MG2Vec approach on the unconstrained metagenome data and present results in Figure~\ref{fig:unconstrained_prc}. We consider 2 other baselines - C2V and GN2V, which use only contextual features (Section~\ref{sec:pre_training}) and only global structural features, respectively, for comparison. As can be seen, the MG2Vec's performance is significantly better compared with the other two baselines, considering that \textit{P. multocida}, \textit{H. somni} and \textit{M. haemolytica} had only 4,980, 545 and 26 reads in their entire metagenomes (over 1 million sequences), respectively. We observed an AUC (Area Under the Curve) of $0.54$, $0.72$ and $0.78$ for MG2Vec for \textit{P. multocida}, \textit{H. somni} and \textit{M. haemolytica}, respectively. The corresponding numbers for GN-2V ($0.52$, $0.73$, and $0.77$) and C2V ($0.52$, $0.74$, and $0.74$) were considerably lower. These numbers indicate that the proposed MG2Vec learns robust features from a smaller, curated number of samples and can help detect pathogen presence in considerably larger, unconstrained metagenome samples. While the numbers are lower than expected, we present one of the first steps to addressing the larger problem of unconstrained, fine-grained pathogen detection from uncurated metagenome samples. 

\subsection{Generalization Evaluation}
Given the promising results from \textit{targeted} pathogen detection, we evaluate our approach on its ability to generalize to other pathogens without having any examples during the training phase. To this end, we setup two evaluation settings: (i) generalization to related but known pathogens and (ii) generalization to completely unknown, out-of-domain pathogens. In the former, we aim to assess the capability to capture representations from metagenome sequences to distinguish between highly similar, yet unknown pathogens in addition to known ones. The latter experiment allows us to observe MG2Vec's ability to rapidly adapt to novel pathogens with limited finetuning efforts. 

\subsubsection{Generalization to Related, Unknown Pathogens}\label{sec:generalDetection}
We extract \textit{MG2Vec} representations from a \textit{complete}, unlabeled, metagenome sample and cluster them to segment the metagenome into groups that comprise samples from each pathogen. The cluster labels are mapped to ground-truth labels using the Hungarian method for quantitative analysis. 
This evaluation allows us to identify the capability of building a general-purpose metagenome analysis pipeline that can enable novel pathogen discovery with minimal supervision. 
We present three different results to assess the generalization capabilities. 

\textbf{Data.} For precise evaluation, we use simulated metagenomes for each of the three cases for evaluation, which allows us to concretely measure how many sequence reads belong to each pathogen class. In the first case, we create a simulated metagenome containing sequences from the Bovine genome along with sequences from the bacteria associated with BRDC as explained in Section~\ref{sec:pathogen_list}. For the second and third cases, we create a simulated metagenome that contains sequences from both known and unknown pathogens (see Section~\ref{sec:pathogen_list}) that are spread across viral and bacterial pathogens beyond those associated with BRDC from the first setting. 

\textbf{Quantitative Results.} First, we see how well a metagenome sequence containing only previously seen, \textit{unlabeled} pathogens is segmented. The confusion matrix is shown in Figure~\ref{fig:pathogen_clustering}(a). It can be seen that pathogens that are distinct from each other i.e., \textit{P. Multocida} and \textit{T. Pyogenes}, are isolated pretty well, whereas there is more confusion among highly related pathogens.  
Note that although some part of the genome of these pathogens were observed during pre-training, they were \textit{unlabeled} and trained in a \textit{species-agnostic} manner, unlike traditional bioinformatics pipelines, which are targeted searches over the entire taxonomy of species. 

Second, we assess the capability of the representations to generalize to \textit{novel} pathogens, including DNA viruses, that were not part of the pre-training pipeline. 
Specifically, the unknown viral pathogens such as \textit{Bovine viral diarrhea virus} (BVDV), \textit{Bovine parainfluenza virus 3} (BPIV-3), \textit{Bovine herpesvirus 1} (BoHV-1), \textit{Bovine coronavirus} (BCoV) and \textit{Bovine respiratory syncytial virus} (BRSV) were unobserved during the pretraining process where only the bacterial pathogens associated with BRDC were present. 
The confusion matrix from this evaluation is shown in Figure~\ref{fig:pathogen_clustering}(b), where a metagenome sample containing three previously \textit{seen} pathogens (\textit{M. haemolytica, T. pyogenes}, and \textit{M. bovis}) and $9$ \textit{unseen} pathogens. It can be seen that the \textit{MG2Vec} representations group the metagenome reads very well and in a manner that suggests that the representations implicitly capture the taxonomic-level relationships between the pathogen species. For example, the two highly related viruses (BHV5 and BHV1), which share $91\%$ of their genome, are confused the most, but not with the more distantly related Bovine Coronavirus (BCV). Similarly, the PI-3 and BRSV viruses belong to the same family \textit{Paramyxoviridae} and the prediction confusion between the two classes is significantly higher. These results suggest that the explicit encoding of contextual and structural properties in the \textit{MG2Vec} representations help implicitly capture hierarchies in the taxonomy and offers an interesting way for the automated analysis of metagenomes. 

Finally, in the third case, we evaluate the capability of the representations to help segment a metagenome sample into groups from different pathogen types (bacteria vs. virus) to accelerate the targeted search process with established bioinformatics pipelines for fine-grained point-of-care diagnosis. We extract the \textit{MG2Vec} representations and cluster the sequence reads into $3$ groups - host, bacteria and virus and present a visualization in Figure~\ref{fig:pathogen_clustering}(c). It can be seen that the representations are robust to segment out host sequences with high precision (F1-score: 0.76) and help distinguish between bacteria and viral sequences with high fidelity, \textit{without training.} 
These results are encouraging signs for integrating deep neural representations with traditional bioinformatics pipelines to speed up the search process, which can take weeks (BLAST) or hours (Kraken2) for a typical clinical sample with $3$ million sequence reads. 

\textbf{Comparison with traditional pipelines.} We also compared our approach's performance on generalized pathogen detection with a traditional bioinformatics pipeline using Kraken2~\cite{wood2019improved}. The standard nucleotide database was used for matching metagenome reads, and Kraken2 returned $3131$ unique detections. For a fair comparison, we filter the detected classes to those found within the metagenome sample and add the extraneous detections to a separate class. This setting is similar to our unsupervised segmentation setting, where the set of clusters is fixed, and the metagenome is clustered into its groups containing genome sequences belonging to the same species. Kraken2 obtains an average F1 score of $0.537$, while our unsupervised segmentation achieves an F1-score of $0.409$ when the number of clusters is set to the actual number of pathogens, and $0.624$ when allowed to over-segment. Note that Kraken2 takes a \textit{segmentation-by-labeling} approach that is prone to raising false alarms. Without binning the erroneous detection, the F1-score is $22.87\%$ which is significantly lower than our unsupervised F1-score of $40.9\%$ when
GT clusters are known. These are encouraging results considering that our representations are learned from metagenomes where these pathogens, particularly the DNA viruses, were unobserved. 
These results show that deep learning representations offer a way forward to isolate novel pathogens with limited labeled data. 
\subsubsection{Generalization to Out-Of-Domain Pathogens}
\begin{table}[t]
    \centering
    \caption{\textbf{Evaluation on Out-of-Domain Data} on a large public dataset with 2505 pathogens. $^*$ indicates BRDC-based features were used and only classifier layer was fine-tuned.}
    \resizebox{0.99\columnwidth}{!}{
    \begin{tabular}{|c|c|c|c|}
    \toprule
        \textbf{Approach} & \textbf{In-Domain Features} & \textbf{Precision} & \textbf{Recall} \\
        \toprule
        CNN-LSTM & \ding{51} & 20.5 & 0.06 \\
        Seq2Species~\cite{busia2019deep} & \ding{51} & 35.7 & 0.22 \\
        CNN & \ding{51} & 70.22 & 1.79\\
        LSTM & \ding{51} & 89.3 & 15.5 \\
        LSTM + Attention~\cite{liang2020deepmicrobes} & \ding{51} & \textbf{94.2} & \textbf{42.8} \\
        \midrule
        \textbf{Ours} (\textit{MG2Vec})$^*$ & \ding{55} & \textbf{91.7} & \textbf{24.8}\\
        \bottomrule
    \end{tabular}
    }
    \label{tab:largeScale_Quant}
\end{table}
In addition to our experiments with the pathogens from the Bovine Respiratory Disease Complex (BRDC), we also evaluate on a large-scale \textit{human} gut microbiome dataset evaluated in DeepMicrobes~\cite{liang2020deepmicrobes}, one of the state-of-the-art approaches in metagenome analysis. This evaluation allows us to assess the generalization capability of the proposed model, pre-trained on BRDC-based microbiome to human gut microbiome with larger number of unrelated pathogens. 

\textbf{Data.} Following prior work~\cite{liang2020deepmicrobes}, we use the large-scale microbiome gut study from the ENA study accession ERP108418~\cite{almeida2019new} to derive $3269$ metagenome-assembled genomes. These genome samples are fragments of the metagenome that are isolated with species-level sequence reads and assembled to form a metagenome file similar to a \textit{16S} sequencing mechanism. The metagenomes were further filtered for quality using CheckM~\cite{parks2015checkm} by selecting genomes with 
$>90\%$ completeness, $0\%$ contamination and strain heterogeneity. The resulting benchmark contained sequences from $2505$ pathogens found in the human gut microbiome.

\textbf{Experimental Setup.} We compare against a variety of deep learning baselines including a convolutional neural network (CNN) with residual connections, a CNN and bidirectional long short-term memory (CNN-LSTM) hybrid model, the seq2species~\cite{busia2019deep} model, and the DeepMicrobes~\cite{liang2020deepmicrobes} baselines - \textit{LSTM} and \textit{LSTM + Attention}. Combined, these baselines represent the state-of-the-art approaches to metagenome analysis. To evaluate our generalization capability, \textbf{we do not train on this out-of-training-domain benchmark}. We only finetune the final classification layer using the labels provided in the training set. This allows us to evaluate the robustness of the proposed approach's features to help identify completely unrelated pathogens with limited training data. 

\textbf{Quantitative Results.} We present the evaluation results in Table~\ref{tab:largeScale_Quant}. It can be seen that our model, \textit{MG2Vec}, can generalize to other domains and outperform many deep learning baselines, all of which are pre-trained and finetuned on the target data. In fact, we significantly outperform Seq2Species, a model designed for short reads and trained for targeted recognition on the human gut microbiome data. We obtain a precision of $91.7\%$ and recall of $24.8\%$ \textit{using out-of-domain features}. This performance is remarkable considering that we do not train on the data (except for the classification layer) and use the features pre-trained on the BRDC data. Note that these numbers are on pair-reads with 16S pair-read sequencing, while our representations are trained on single read shotgun metagenome sequencing and hence demonstrates the effectiveness and scalability of the learned representations to novel species, sequencing approaches and tasks.
Experiments in Section~\ref{sec:generalDetection} were intended to showcase the effectiveness of our learned representations to segment novel, \textit{related} pathogens that were not in the pre-training data, whereas these experiments show the effectiveness of our representations to generalize to completely unknown pathogen sequences. Combined, these two experiments demonstrate the generalized nature of the proposed framework and hence forms the first steps towards a species-agnostic representation learning framework that can learn to identify new classes with limited supervision.

\subsection{Ablation Studies}\label{sec:ablation}
We also systematically evaluate the contributions of the different components of the proposed approach and present each evaluation in detail below. 

\begin{table}[t]
    \centering
    \caption{\textbf{Quality of intermediate representations.} Performance of various machine learning baselines using intermediate representations from our framework. Average F1 scores are reported.
    }
    \resizebox{0.99\columnwidth}{!}{
     \begin{tabular}{|c|c|c|c|c|c|c|c|c|}
    \toprule
    \multirow{2}{*}{\textbf{Approach}} & \multicolumn{2}{|c|}{\textbf{G-N2V}} & \multicolumn{2}{|c|}{\textbf{C2V}} & \multicolumn{2}{|c|}{\textbf{Encoder}} & \multicolumn{2}{|c|}{\textbf{MG2V}}\\
    \cline{2-9}
    & \textbf{Host} & \textbf{Path.} & \textbf{Host} & \textbf{Path.} & \textbf{Host} & \textbf{Path.} & \textbf{Host} & \textbf{Path.} \\
    \toprule
    LR       & 0.979  & 0.515  & 0.979  & 0.526  & 0.915 & 0.233  &  0.979  & 0.509 \\
    SVM      & 0.975  & 0.525  &  0.979 & 0.536  & 0.904 & 0.171  &  0.975  & 0.405 \\
    MLP      & \textbf{0.985}  & 0.542  & 0.979  & 0.543  & 0.935 & 0.369  &  \textbf{0.985}  & 0.534 \\
    DL       & 0.979  & 0.565  & 0.979  & 0.567  & 0.861  & 0.276   & 0.981  & \textbf{0.631}\\
    \midrule
    K-Means  & 0.426  & 0.225  & 0.471  &  0.245  & 0.358  & 0.144  & 0.441  & 0.368 \\
    \bottomrule
    \end{tabular}
    }
    
    \label{tab:rep_ablation}
\end{table}
\textbf{Intermediate Representations.} We first evaluate the effectiveness of the different intermediate representations that can be obtained from our approach and summarize the results in Table ~\ref{tab:rep_ablation}. We consider four different intermediate representations - only the global structural priors from Section~\ref{sec:globalStructure} (G-N2V), only the contextualized representations from the transformer network's embedding layer from Section~\ref{sec:pre_training} (C2V), the output of the transformer encoder network $H_i$ from Section~\ref{sec:pre_training} (Encoder) and our final \textit{MG2Vec} representation (MG2V). It can be seen that all representations help distinguish between host and pathogen, as indicated by the high F1 score for the host class. However, they do not help distinguish between the fine-grained classes within the pathogens. The structural priors by themselves do well considering that it does not take any pre-training or significant computation time (for smaller k-mers), while the encoder outputs surprising do not capture the fine-grained differences among pathogen classes. 

\begin{table}[t]
    \centering
    \caption{\textbf{Ablation Study:} Evaluation of design choices in the proposed approach such as \textit{k-mers} and structural priors. Precision (\textit{Prec.}), Recall and F1 scores \textit{averaged across classes} are reported. }
    \resizebox{0.99\columnwidth}{!}{
    \begin{tabular}{|c|c|c|c|}
    \toprule
    \textbf{Approach} & \textbf{Prec.} & \textbf{Recall} & \textbf{F1}\\
    \midrule
    \multicolumn{4}{|c|}{\textit{Effect of K-Mers}}\\
    \midrule
    $k=3$ & 0.61 & 0.66 & 0.634\\
    \textbf{$k=4$} & \textbf{0.63} & \textbf{0.72} & \textbf{0.67}\\
    $k=6$ & 0.63 & 0.66 & 0.65\\
    \midrule
    \multicolumn{4}{|c|}{\textit{Effect of Structural Priors}}\\
    \midrule
    without Global Prior  & 0.61 & 0.67 & 0.64 \\
    without Normalized Weights & 0.56 & 0.65 & 0.60 \\
    with Unidirectional Attention  & 0.60 & 0.67 & 0.63 \\
    \midrule
    \multicolumn{4}{|c|}{\textit{Larger k-mers without pretraining}}\\
    \midrule
    $k=10$ & 0.49 & 0.63 &  0.55\\
    $k=12$ & 0.50 & 0.64 & 0.56 \\
    \midrule
    \multicolumn{4}{|c|}{\textit{Effect of contextualization networks}}\\
    \midrule
    1-layer LSTM & {0.60} & {0.51} & {0.54}\\
    Bi-LSTM & 0.54 & 0.52 & 0.52\\
    Transformer & \textbf{0.63} & \textbf{0.72} & \textbf{0.67}\\
    \bottomrule
    \end{tabular}
    }
    \label{tab:ablation}
\end{table}

\textbf{Effect of Structural Priors.} We also evaluate the effect of structural priors on the \textit{MG2Vec} representations by changing how they are constructed. We consider three different settings - no global priors to initialize the embeddings (Section~\ref{sec:globalStructure}), without normalizing the edge weights in the global graph structure (Equation~\ref{eqn:weight_update}) and with only unidirectional attention, i.e., the transformer is trained as an autoregressive model. It can be seen from Table~\ref{tab:ablation} that the design with the highest impact is the normalization of the edge weights in the global structural graph and could be attributed to the fact that the biased random walks for capturing the global structure would be highly impacted by spurious patterns that can occur due to observation noise. Interestingly, all alternative design choices have significantly lower recall (particularly among pathogen classes), which indicates that the use of global priors and bidirectional attention help capture differences between similar reads.

\textbf{Effect of K-Mers}. We also vary the length of each k-mer to identify the optimal structure. Note that we only consider $k{=}3\ldots6$ for the entire pipeline due to scalability issues since $k{>}8$ leads to a rather large vocabulary (more than a million) and hence does not converge. We consider larger k-mers ($k{=}10,12$) with only the global priors and summarize the results in  Table~\ref{tab:ablation}, where it can be seen that $k{=}4$ provides the best results, corroborating prior studies showing similarity in tetranucleotide bases between closely related species~\cite{perry2010distinguishing}. 

\textbf{Effect of contextualization network.} Finally, we evaluate the impact of the transformer architecture on the contextualization process (~\ref{sec:pre_training}) by varying the network architecture of the model to instead use LSTM and Bi-LSTM models. These models can be considered to be analogous to architectures present in other baselines such as DeepMicrobes~\cite{liang2020deepmicrobes} and Seq2Species~\cite{busia2019deep}. As can be seen from Table~\ref{tab:ablation}, the transformer architecture demonstrates a better tendency to capture the intra-class variations better between the pathogens and hence provides a better F1-score across pathogen and host classes. 

\section{Discussion and Future Directions}\label{sec:conclusion}
In this work, we presented \textit{MG2Vec}, a deep neural representation that explicitly captures the global structural properties of the metagenome sequence and produces \textit{contextualized} representations of k-mers. With extensive experiments, we evaluate and demonstrate the effectiveness of using such representations for both targeted and generalized pathogen detection for scalable metagenome analysis. We show that the proposed representation offers competitive performance to traditional bioinformatics pipelines for segmenting metagenome samples into genome reads from different pathogen species with limited labeled data. 
We draw inspiration from these results and aim to integrate deep neural representations into traditional bioinformatics pipelines to accelerate the detection of novel pathogens.
\section{Acknowledgements}
This research was supported in part by the US Department of Agriculture (USDA) grants AP20VSD and B000C011.

We thank Dr. Kitty Cardwell and Dr. Andres Espindola (Institute of Biosecurity and Microbial Forensics, Oklahoma State University) for providing access and assisting with use of the MiFi platform. We also thank Ms. Vineela Indla and Ms. Vennela Indla for their discussions on machine learning baselines.
\bibliography{egbib}
\bibliographystyle{IEEEtran}

%

\begin{IEEEbiographynophoto}{Sai Narayanan} received B.V.Sc. \& A.H., M.V.Sc (Veterinary Clinical Medicine) and the Post-graduate Diploma in Zoonosis from Madras Veterinary College, Tamil Nadu Veterinary and Animal Sciences University in Chennai, India. He is currently pursuing a Ph.D. in Veterinary Biomedical Sciences from the College of Veterinary Medicine at Oklahoma State University. His research focus include development of diagnostics, next generation sequencing applications in diagnostics, metagenomics and microbiome assessment.
\end{IEEEbiographynophoto}
\vspace{-0.2in}

\begin{IEEEbiographynophoto}{Sathyanarayanan N. Aakur}
received the B.Eng. degree in Electronics and Communication Engineering from Anna University, Chennai, India in 2013. He received the M.S. degree in Management Information Systems and the Ph.D. degree in Computer Science from the University of South Florida, Tampa, in 2015 and 2019, respectively. He is currently an Assistant Professor with the Department of Computer Science at Oklahoma State University since 2019. His research interests include self-supervised learning, commonsense reasoning for visual understanding, and deep learning applications for genomics. He is the recipient of the National Science Foundation CAREER award in 2022 and has served as Associate Editor for IEEE Robotics and Automation Letters since 2021.
\end{IEEEbiographynophoto}

\begin{IEEEbiographynophoto}{Priyadharsini Ramamurthy} received the B.Eng. degree in Computer Science from Anna University, Chennai, India in 2006. She received the M.S. degree in Engineering Management from International Technological University and is currently pursuing her Ph.D. degree in Computer Science from Oklahoma State University from 2019. Her research interests include Natural Language Understanding, Intelligent Agents, multi-model learning, and deep learning applications.
\end{IEEEbiographynophoto}

\begin{IEEEbiographynophoto}{Arunkumar Bagavathi} is an Assistant Professor in the Department of Computer Science at OklahomaState University. He received his B.Eng. degree in Computer Science from Anna University, India in 2014 and he received his Ph.D. degree in Computer Sceince from the University of North Carolina at Charlotte in 2019. His research interests include data mining, network science, computational social science, and applied machine learning. His research works have been published in venues like Complex Networks, IEEE/ACM ASONAM, IEEE ICMLA, and JMIR.

\end{IEEEbiographynophoto}

\begin{IEEEbiographynophoto}{Vishalini Ramnath} received the B.Eng. degree in Electronics and Communication Engineering from Anna University, Chennai, India in 2013. She received the M.S. degree in Computer Science from the University of Illinois, Chicago and the Ph.D. degree in Computer Science and Engineering from the University of South Florida, Tampa, in 2016 and 2020, respectively. She is currently a Teaching Assistant Professor with the Department of Computer Science at Oklahoma State University since 2019. Her research interests include machine learning for the Internet of Things, machine learning for hardware security, and deep learning applications for genomics.
\end{IEEEbiographynophoto}
\vfill
\begin{IEEEbiographynophoto}{Akhilesh Ramachandran} received the B.V.Sc. \& A.H. degree from Kerala Agricultural University, Kerala, India in 1998. He received the Ph.D. in Veterinary Biomedical Sciences in 2003 from Oklahoma State University. He is currently an Associate Professor with the Department of Veterinary Pathobiology at Oklahoma State University.  He also serves as the Head of the Microbiology and Molecular Diagnostics Sections at the Oklahoma Animal Disease Diagnostic Laboratory. His research interests include development of novel molecular diagnostic protocols and platforms for infectious disease diagnosis.  
\end{IEEEbiographynophoto}




\end{document}